\documentclass{article}

\usepackage{microtype}
\usepackage{graphicx}
\usepackage{caption}
\usepackage{subcaption}
\usepackage{tabularray}
\usepackage[table,xcdraw]{xcolor}
\usepackage{soul}
\usepackage{colortbl}
\usepackage{listings}
\usepackage[all, defaultlines=3]{nowidow}

\usepackage{hyperref}

\usepackage[accepted]{mlsys2025}

\mlsystitlerunning{EdgeRunner 20B}

\begin{document}

\twocolumn[
\mlsystitle{EdgeRunner 20B:\\Military Task Parity with GPT-5 while Running on the Edge}

\mlsyssetsymbol{equal}{*}

\begin{mlsysauthorlist}
\mlsysauthor{Jack FitzGerald}{equal,edge}
\mlsysauthor{Aristotelis Lazaridis}{equal,edge}
\mlsysauthor{Dylan Bates}{equal,edge}
\mlsysauthor{Aman Sharma}{equal,edge}
\mlsysauthor{Jonnathan Castillo}{edge}
\mlsysauthor{Yousif Azami}{edge}
\mlsysauthor{Sean Bailey}{edge}
\mlsysauthor{Jeremy Cao}{edge}
\mlsysauthor{Peter Damianov}{edge}
\mlsysauthor{Kevin de Haan}{edge}
\mlsysauthor{Luke Kerbs}{edge}
\mlsysauthor{Vincent Lu}{edge}
\mlsysauthor{Joseph Madigan}{edge}
\mlsysauthor{Jeremy McLaurin}{edge}
\mlsysauthor{Jonathan Tainer}{edge}
\mlsysauthor{Dave Anderson}{edge}
\mlsysauthor{Jonathan Beck}{edge}
\mlsysauthor{Jamie Cuticello}{edge}
\mlsysauthor{Colton Malkerson}{edge}
\mlsysauthor{Tyler Saltsman}{edge}
\end{mlsysauthorlist}

\mlsysaffiliation{edge}{EdgeRunner AI}

\mlsyscorrespondingauthor{Jack FitzGerald}{research@edgerunnerai.com}

\mlsyskeywords{Machine Learning, MLSys}

\vskip 0.3in

\begin{abstract}
\vspace{3pt}
We present EdgeRunner 20B, a fine-tuned version of gpt-oss-20b optimized for military tasks. EdgeRunner 20B was trained on 1.6M high-quality records curated from military documentation and websites. We also present four new tests sets: (a) combat arms, (b) combat medic, (c) cyber operations, and (d) mil-bench-5k (general military knowledge). On these military test sets, EdgeRunner 20B matches or exceeds GPT-5 task performance with 95\%+ statistical significance, except for the high reasoning setting on the combat medic test set and the low reasoning setting on the mil-bench-5k test set. Versus gpt-oss-20b, there is no statistically-significant regression on general-purpose benchmarks like ARC-C, GPQA Diamond, GSM8k, IFEval, MMLU Pro, or TruthfulQA, except for GSM8k in the low reasoning setting. We also present analyses on hyperparameter settings, cost, and throughput. These findings show that small, locally-hosted models are ideal solutions for data-sensitive operations such as in the military domain, allowing for deployment in air-gapped edge devices.
\end{abstract}
]

\printAffiliationsAndNotice{\mlsysEqualContribution} %

\raggedbottom
\section{Introduction}

Military technology and infrastructure must have redundancy and resiliency \cite{jakovich2024quantifying}. In private-sector commercial applications, there are many advantages of a cloud-based serving architecture for foundation models: Requests can be consolidated and batched to make maximum use of the inference hardware, traffic patterns can be analyzed for further optimization, and the compute on the edge device can be reduced, thereby lowering cost for the end user. Wartime scenarios, however, can and do remove the underlying assumptions of this architecture. First, either the data center or the network connecting the user to the data center can be destroyed or degraded. Second, there are security restrictions (including in peacetime) that preclude the use of public networks, particularly for classified data and use cases. For these reasons and more, edge computing remains essential for the military.

One may assume that smaller, edge-based foundation models would be inferior to their cloud-based counterparts in terms of task performance. Though holistically this may be true, in this work we show that specialized edge models can achieve parity with cloud offerings for a range of military-specific use cases. The edge model may not match a cloud model for things like writing a paragraph of prose that mimics Elizabethan literature, but it can match and even exceed cloud-based task performance on use cases that warfighters depend on, including battlefield medicine, combat arms, vehicle maintenance, or synthesis of military plans.

\begin{figure*}
    \centering
    \begin{subfigure}[b]{0.49\textwidth}
       \centering
       \includegraphics[width=\textwidth]{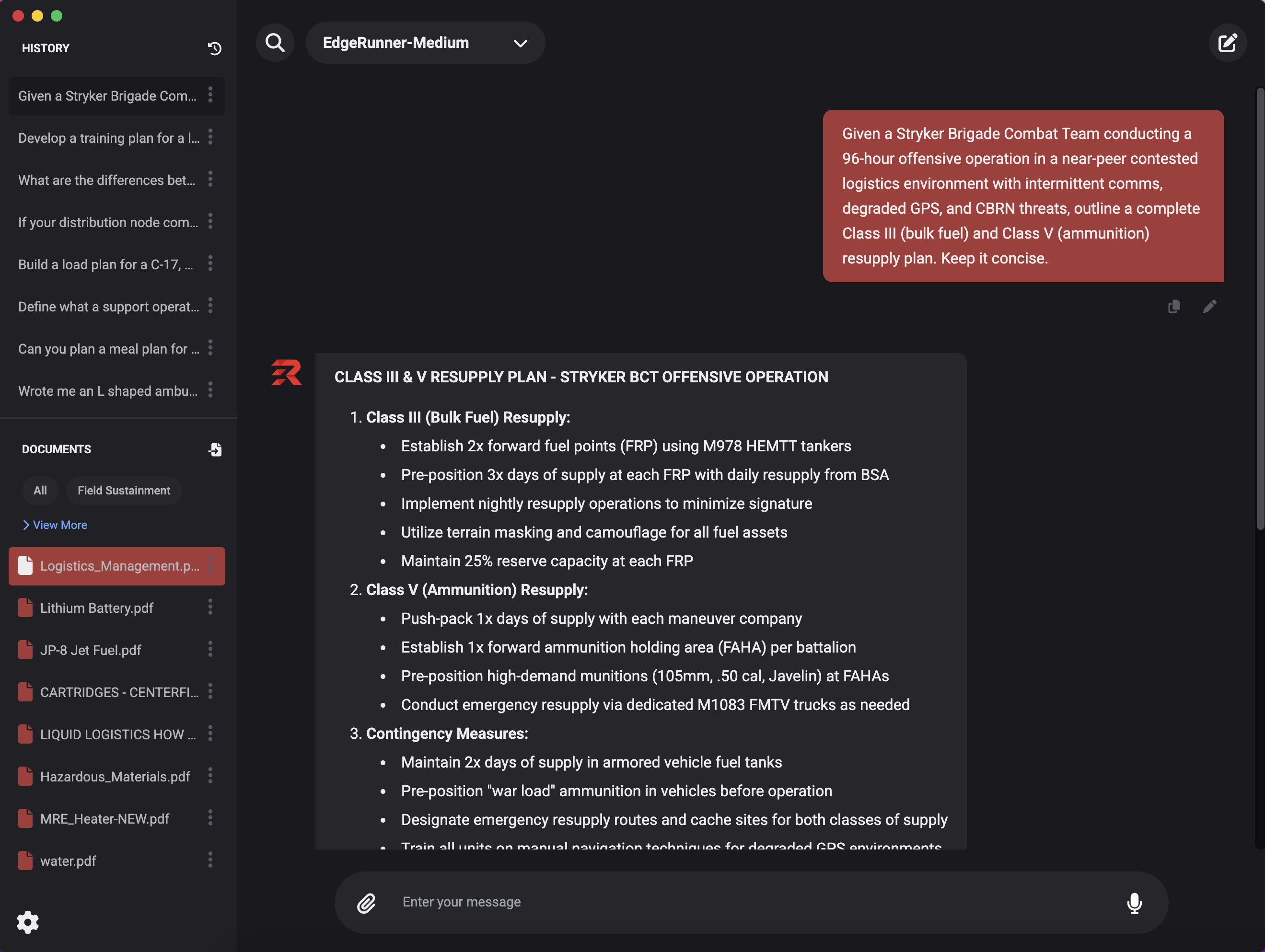}
       \caption{A query given after selecting a document for RAG.}
    \end{subfigure}
    \begin{subfigure}[b]{0.49\textwidth}
       \centering
       \includegraphics[width=\textwidth]{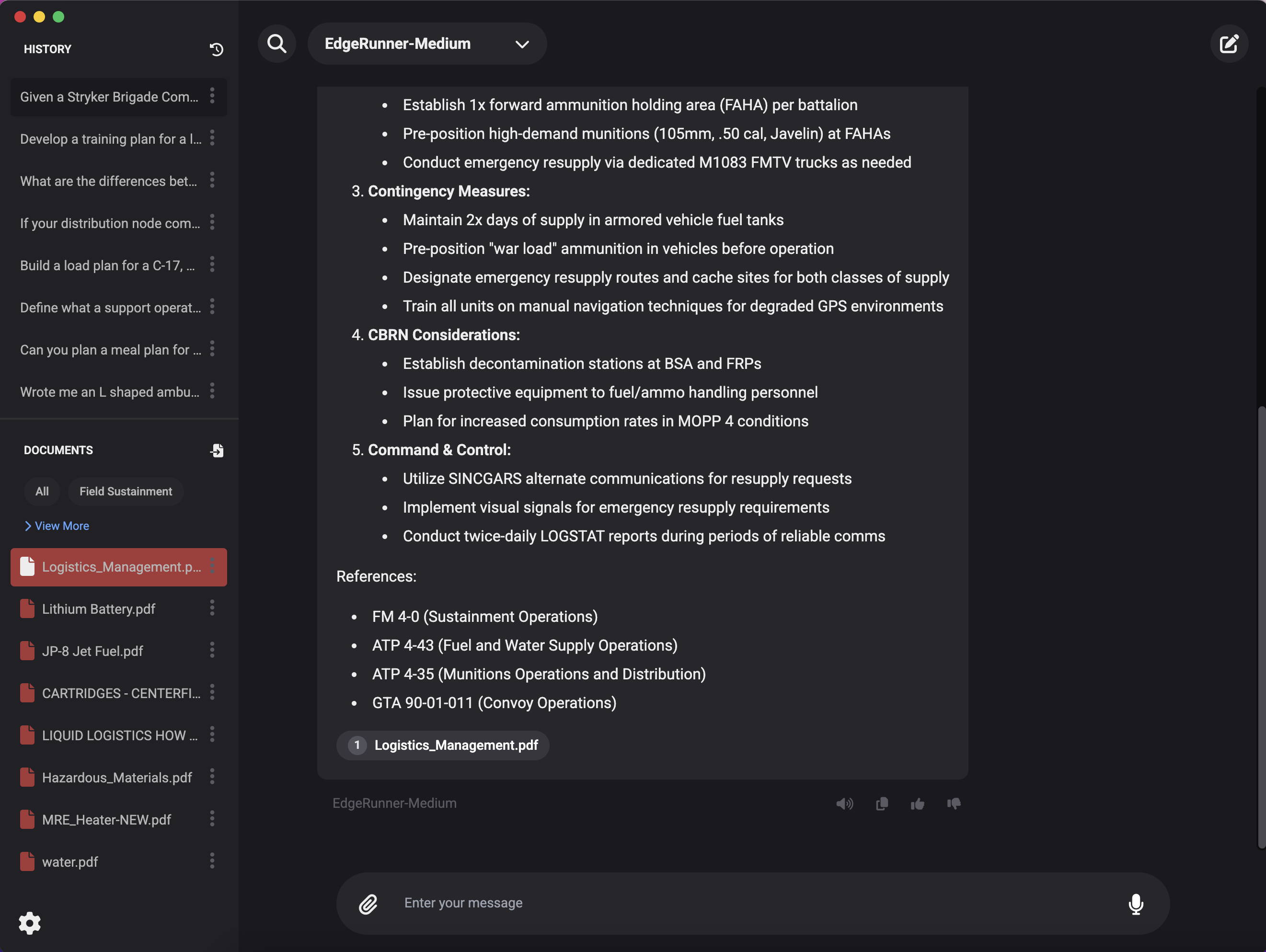}
       \caption{The answer to the query and the citation link(s).}
    \end{subfigure}
    \\[2ex]
    \begin{subfigure}[b]{0.49\textwidth}
       \centering
       \includegraphics[width=\textwidth]{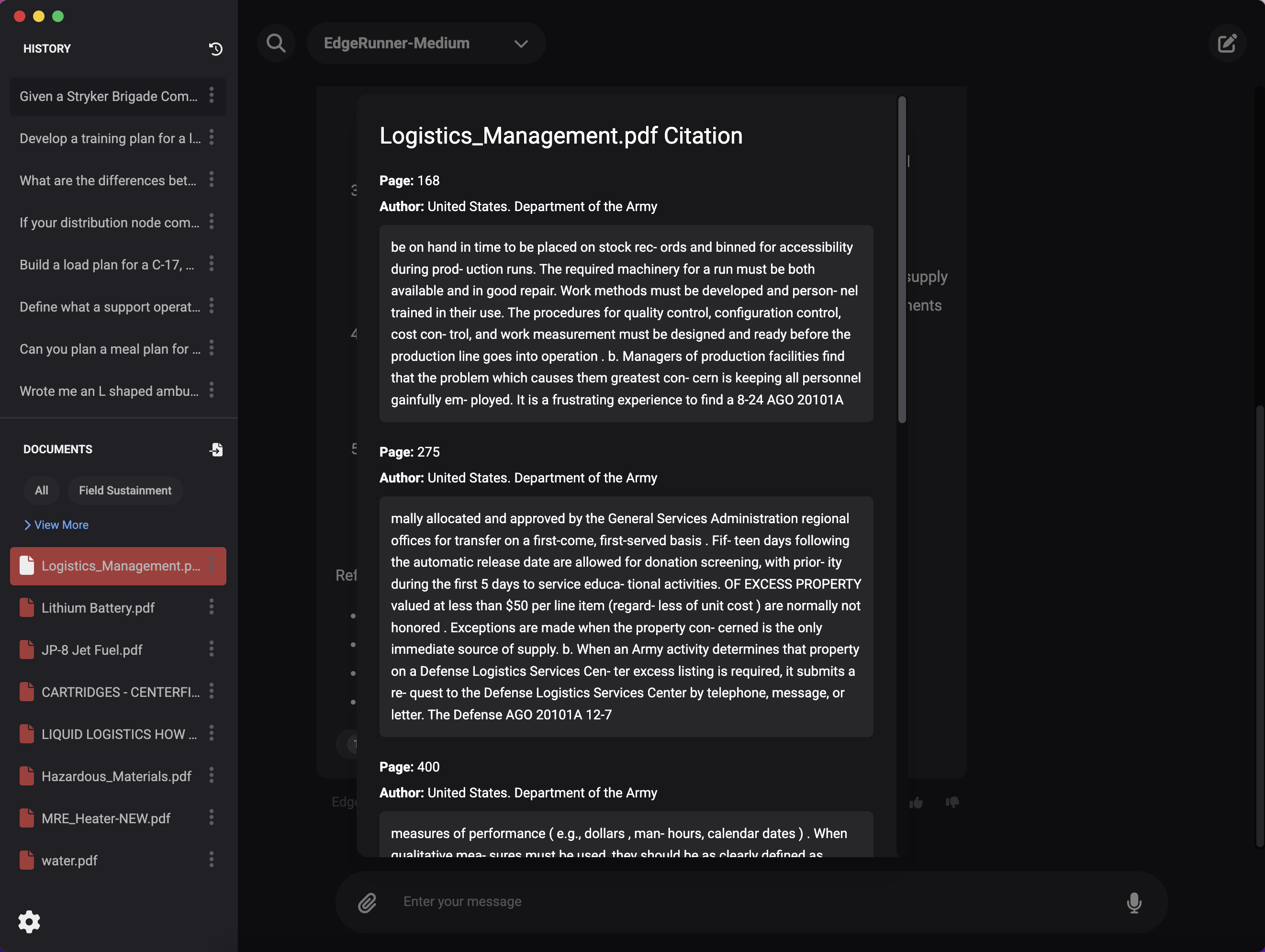}
       \caption{RAG references used by the model.}
    \end{subfigure}
    \begin{subfigure}[b]{0.49\textwidth}
       \centering
       \includegraphics[width=\textwidth]{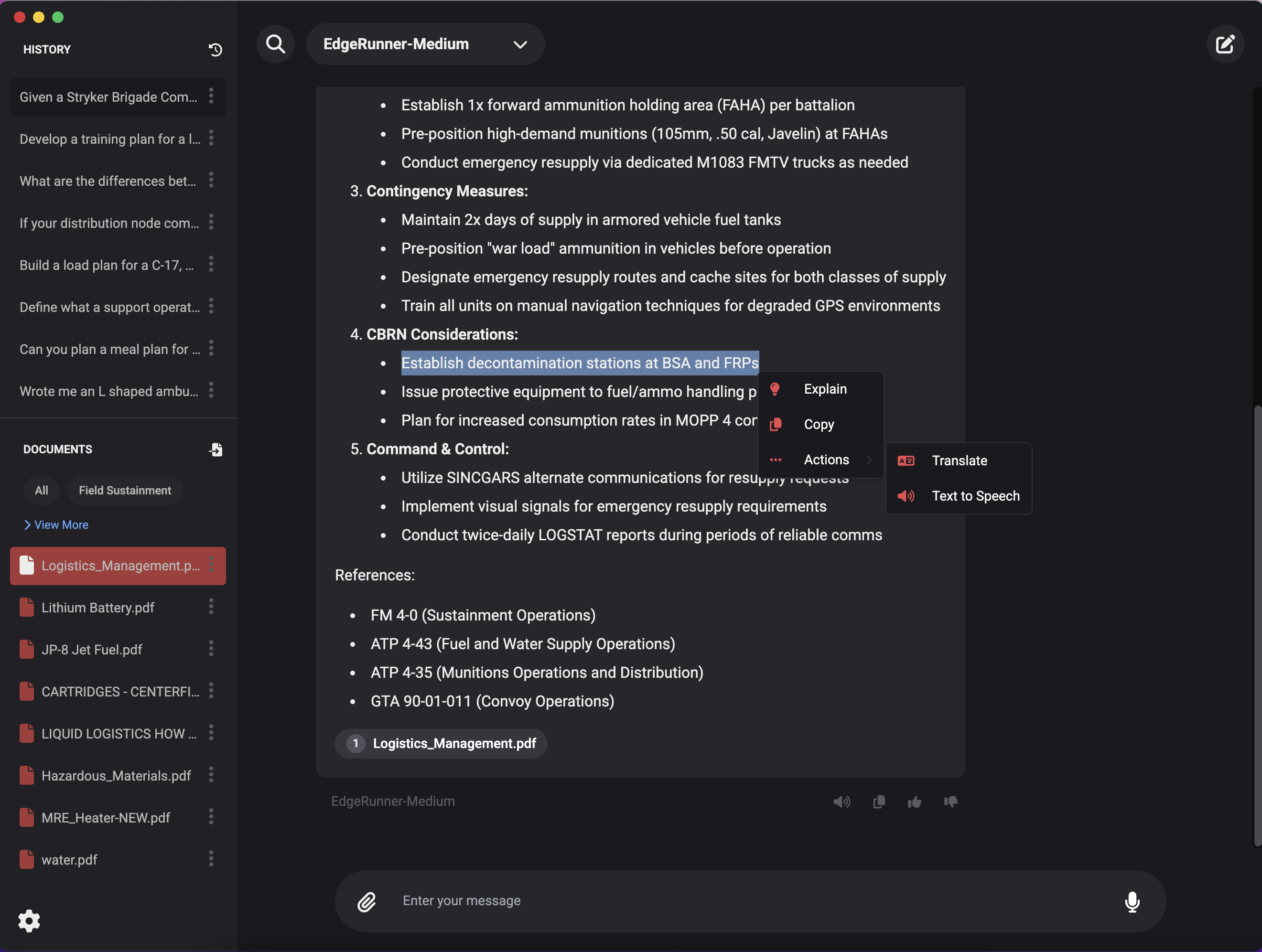}
       \caption{Explanation, Editing, Speech, and Translation features.}
    \end{subfigure}
    \\[2ex]
    \begin{subfigure}[b]{0.49\textwidth}
       \centering
       \includegraphics[width=\textwidth]{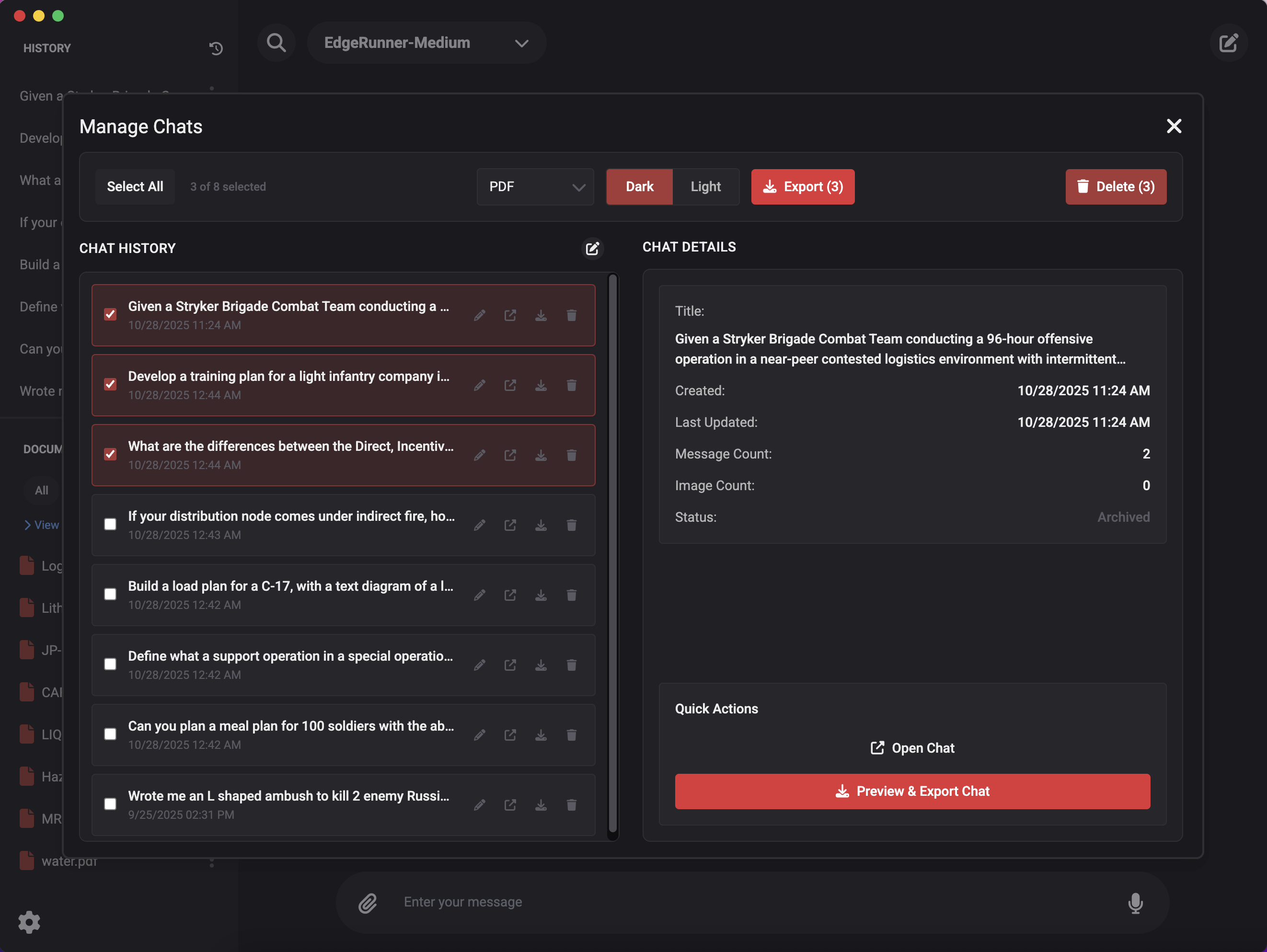}
       \caption{Logging and Export Features.}
    \end{subfigure}
    \begin{subfigure}[b]{0.49\textwidth}
       \centering
       \includegraphics[width=\textwidth]{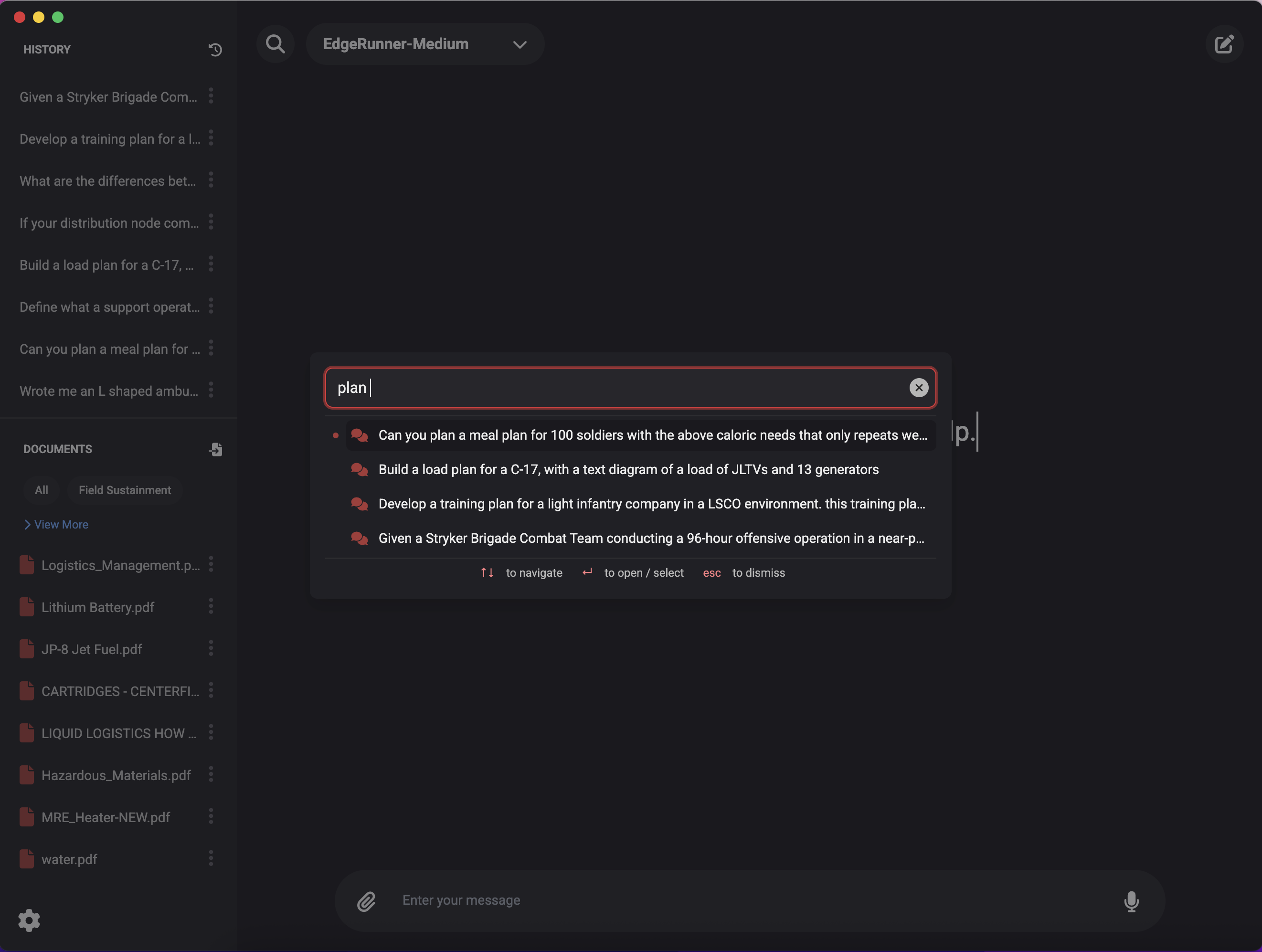}
       \caption{Search features.}
    \end{subfigure}
\caption{The EdgeRunner inference system for running military-specific LLMs locally on the user's device and performing Retrieval Augmented Generation (RAG).}
\label{fig:edgerunner-interface}
\end{figure*}

EdgeRunner is a product and system for deploying foundation models to edge devices. It supports a wide range of model sizes, modalities, and capabilities across hardware ranging from local server deployments and powerful laptops to commodity laptops and mobile phones. An example of the EdgeRunner interface is given in Figure \ref{fig:edgerunner-interface}.

Though the EdgeRunner ecosystem is broad, in this paper we've targeted a single pair of models, being gpt-oss-20b \cite{openai2025gptoss120bgptoss20bmodel}, which was released by OpenAI to the public and was further trained and tuned by us, and GPT-5 \cite{gpt5-system-card}, which is cloud-based, closed, and at the time of this writing is widely regarded as one of the top foundation models in the world. Our contributions are as follows:

\begin{itemize}
    \item We demonstrate that a specialized edge model can be trained to attain military task performance parity with its premier, cloud-based counterpart,
    \item We introduce and analyze new military-specific test sets that we have developed, and
    \item We provide inference and cost benchmarking.
\end{itemize}

\section{Test Sets}

\begin{table*}[tbh]
\centering
\begin{tblr}{
    colspec={lcccX[l]},
    rows={m},
    row{1} = {font=\bfseries}
}
\hline[1pt]
Name & Classification & Size & Epochs & Description \\
\hline[0.5pt]
\textsc{combat-arms} & Silver & 180 & 3 & A dataset covering the combat arms career field with a focus on infantry-related Q\&A pairs. The dataset was vetted by three US Army officers with combined experience of over 45 years, including over 20 years of special missions experience. \\
\textsc{combat-medic} & Silver & 446 & 2 & A dataset covering the combat medic career field. The dataset was vetted by three US Army officers with combined experience of over 45 years, including over 20 years of special missions experience. \\
\textsc{cyber} & Gold & 142 & 3 & A from-scratch cyber security and operations dataset written by a former US Army cyber expert covering (a) compliance, (b) training, (c) incident response, (d) mission planning, (e) security procedures, (f) threat intelligence, and (g) tooling. \\
\textsc{mil-bench-5k} & Silver & 5,000 & 1 &  A general-purpose military dataset covering many topics. The source documents used for this task were publications across the Army, Joint Staff, DoD and Logistics domains, targeting roughly equal representation for each domain within the benchmark. The dataset was vetted by three US Army officers with combined experience of over 45 years, including over 20 years of special missions experience. \\
\hline[1pt]
\end{tblr}
\caption[lorem]{Military-specific test sets used in this work. Epochs are the number of repetitions (with differing seed) used when evaluating.}
\label{tab:mil-test-sets}
\end{table*}

\subsection{Military-Specific Test Sets}
Our military test sets consist of Question and Answer (Q\&A) pairs each with a realistic question a user might ask a model on a given topic and a correct answer based on official doctrinal sources.

We consider three major classes of test sets:
\begin{enumerate}
    \item \textbf{Gold test sets}, in which Subject Matter Experts (SMEs) create every input and output in the dataset from scratch and use no AI assistance,
    \item \textbf{Silver test sets}, in which SMEs verify the inputs and outputs for accuracy, relevance, and realism, but AI assistance is used in some way, whether to create candidate input data, to curate or filter possible inputs, or to assist in authoring the outputs, and
    \item \textbf{Bronze test sets}, which were created by our research team but have not been reviewed by military SMEs.
\end{enumerate}

Our silver and bronze datasets are the result of our synthetic data generation pipeline (Cf. Section \ref{sect:training-data}), repurposed to generate evaluation data instead of training data.
Even though the prompts used for generating the silver and bronze benchmarks were substantially different from those used for generating the training dataset(s), we took further measures to ensure that the train set was not polluted by the benchmark's samples (or vice versa) after they were created using text analytics. Specifically, we performed exact-match search on the individual ``question'' and ``answer'' fields of both sets, and as a second pass, we performed fuzzy match search in order to ensure that there were no samples with high-degree overlaps. In the rare cases that such samples were found, they were removed.

Though silver test sets are fully vetted by the SME, some bias may still remain due to the use of AI assistance. Gold test sets, on the other hand, are substantially more expensive to create. Bronze test sets are the cheapest and most scalable to create, but their fidelity to the needs of the warfighter is, by definition, unvetted. We currently use bronze sets only for development and do not report their results here.

A description of our military-specific test sets is provided in Table \ref{tab:mil-test-sets}. We currently keep our test sets hidden for reasons of national security, but any US military member or civil servant is welcome to contact the corresponding author to request an audit of the data and our evaluation methods.

\subsection{General-Purpose Test Sets}

To measure possible regressions in general-purpose capabilities, we use the following public test sets, which we chose to cover a wide breadth of model use cases:

\begin{itemize}
    \item \textbf{ARC-Challenge} \cite{clark2018thinksolvedquestionanswering}: A multiple-choice dataset of grade-school science questions.
    \item \textbf{GPQA Diamond} \cite{rein2023gpqagraduatelevelgoogleproofqa}: The most difficult split (diamond) from the multiple-choice Google-Proof Q\&A benchmark, which consists of very difficult questions for which PhD candidates only had a 65\% accuracy rate at time of publication.
    \item \textbf{GSM8k} \cite{cobbe2021trainingverifierssolvemath}: A multiple-choice dataset of 8k examples of Grade School Math (GSM).
    \item \textbf{IFEval} \cite{zhou2023instructionfollowingevaluationlargelanguage}: An instruction-following dataset consisting of easy-to-verify instructions, such as mentioning a keyword a certain number of times or writing a response with more than a certain number of words.
    \item \textbf{MMLU Pro} \cite{wang2024mmluprorobustchallengingmultitask}: An updated version of the popular Massive Multitask Language Understanding (MMLU) multiple-choice benchmark \cite{hendryckstest2021} designed to be more difficult and more robust to variations in the prompt template.
    \item \textbf{TruthfulQA} \cite{lin-etal-2022-truthfulqa}: A multiple-choice dataset based on popular misconceptions across 38 categories, including health, law, finance and politics.
\end{itemize}

\section{Methods}

\subsection{Training Data}
\label{sect:training-data}

Our training data is synthesized through a multi-stage question-answer generation pipeline that transforms domain-specific documents into high-quality instruction-tuning datasets. Initially, the pipeline digests the input documents, performs text extraction if necessary, and organizes them into manageable chunks for parallel processing. These chunks are then formatted into batch inference requests and processed through LLMs using vLLM as the inference engine \cite{kwon2023efficient}. At each stage the language models are prompted to perform each necessary task using domain-specific instructions, which were derived after multiple iterations of subsequent model training runs on the resulting training data, along with the respective evaluations.

In the first stage of the pipeline, a language model is responsible for generating a high-quality summary of each document, which is used in the next stage (Q-A generation) as context to the isolated chunks of that document. We found that this approach allows the model to provide more relevant and accurate Q-A pairs than using only the input chunks in isolation.

In the second stage, a different language model is prompted to generate multiple diverse question-answer pairs that cover the key concepts, procedures, and details within each document. Raw model outputs undergo initial cleaning to parse the generated Q-A pairs, filter out uninformative responses, and recombine multi-part documents that were split.

In the third and final stage, the pipeline incorporates rigorous quality control, in which each generated Q-A pair is evaluated by a different language model using quality assessment prompts. This evaluator categorizes pairs into three classes: high-quality pairs that pass without modification (PASS), pairs with minor issues (e.g. formatting) that can be corrected through automated rewrites (FIX), and pairs that fail quality standards and are discarded (FAIL). For pairs marked as FIX, the model generates the fixed versions. The final dataset combines both PASS and successfully FIXED pairs, ensuring that the training data meets consistent quality standards for clarity, doctrinal accuracy, completeness, and relevance.

This automated pipeline enables scalable generation of domain-specific instruction-tuning data while maintaining quality through multi-stage filtering and refinement, producing datasets that preserve the technical precision and specificity of the source material. In this paper we use a final dataset of 1.6M records.

\subsection{Training Setup}

Our training infrastructure leverages a containerized, multi-node distributed training pipeline built on SLURM workload management \cite{slurm}. We experimented with TRL \cite{vonwerra2022trl}, Unsloth \cite{unsloth}, the Quantization Aware Training feature in TensorRT Model Optimizer \cite{modelopt}, and Axolotl \cite{axolotl}. All results in this paper are based on runs using Axolotl. The pipeline consists of integrated components that orchestrate experiment setup and distributed execution, using Enroot containers for reproducible software environments. Training is launched as a SLURM job with user-specified resource requirements. We use PyTorch's distributed launcher with the c10d rendezvous backend, enabling fault-tolerant multi-node synchronization across all allocated GPUs, as well as with appropriate NCCL settings for InfiniBand communication. The pipeline configures per-rank Triton and TorchInductor cache isolation to prevent race conditions in kernel compilation during distributed execution.

gpt-oss-20b was released using MXFP4 quantization in the expert layers. Though we experimented with quantization-aware training, all results presented in this paper were based on upcasted bfloat16 training followed by post-training quantization to MXFP4. The post-training quantization was conducted using the \texttt{convert\_oai\_mxfp4\_weight\_only.py} script in Nvidia's TensorRT Model Optimizer library.

\subsection{Evaluation Setup}

For all evaluations in this paper, we used Inspect, a framework for large language model evaluations created by the UK AI Security Institute \cite{UK_AI_Security_Institute_Inspect_AI_Framework_2024}. We used version \texttt{0.3.130} from Sep 6th, 2025.

All military test sets produce freeform outputs that are graded with a judge model. In this work we used the W8A8 quantized version of the Atla Selene 1 model \cite{alexandru2025atlaseleneminigeneral}, which is based on Llama 3.3 70B \cite{grattafiori2024llama3herdmodels} and is further trained using Supervised Fine Tuning (SFT) and Direct Preference Optimization (DPO) to be a better judge. To remove additional variables, we use Inspect's default prompt for the LLM judge, and we do not use any system or developer prompt for any of the military tasks. Further details on the evaluation setup are provided in Appendix \ref{sect:app-eval-setup}.

The general-purpose test sets were evaluated using the task definitions provided in the \texttt{inspect\_evals} library \cite{inspect_evals}, commit \texttt{a27622c}. All general-purpose tasks were graded using heuristics like multiple-choice or other verification, meaning that no model judge was required.

\section{Results and Analysis}

\subsection{Fine Tuning Accuracy}

\begin{figure*}[htbp]
    \centering
    \includegraphics[width=0.9\linewidth]{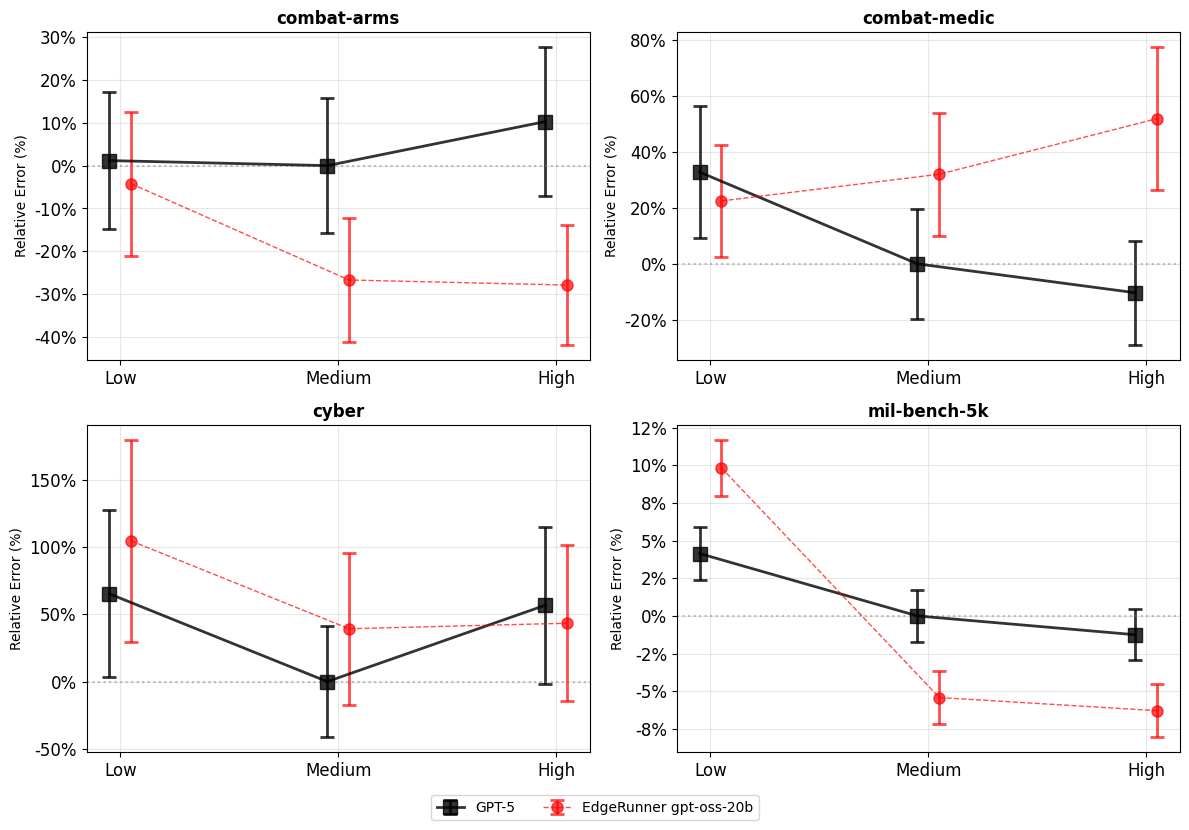}
    \caption{A visual representation of the data presented in Table \ref{tab:mil-results} showing the relative error of EdgeRunner's fine tuned gpt-oss-20b model relative to GPT-5 with medium reasoning. Error bars are given for the standard error. Lower numbers are better (less error).}
    \label{fig:mil-results}
\end{figure*}

Results from gpt-oss-20b fine tuning are provided in Tables \ref{tab:mil-results} and \ref{tab:mil-p-values} and Figure \ref{fig:mil-results}, including for each of the four test sets and for all three inference reasoning levels. Error rates are given relative to medium reasoning with GPT-5, and p-values are also calculated based on the original, occluded absolute scores. 

\begin{table*}[htbp]
\caption[]{Relative error of the EdgeRunner 20B fine-tuned version of gpt-oss-20b versus the medium reasoning setting of GPT-5 (\texttt{gpt-5-2025-08-07}). Lower numbers are better (less error), and negative numbers mean less error than the medium reasoning setting of GPT-5. Standard error is also provided.}
\label{tab:mil-results}
\vspace{3pt}
\centering
\small
\begin{tblr}{
    colspec={X[-1,c]X[-1,c]X[-1,c]X[-1,c]X[-1,c]X[-1,c]X[-1,c]X[-1,c]X[-1,c]X[-1,c]},
    rows={m},
    column{1} = {font=\scriptsize},
    column{2} = {font=\scriptsize},
    column{4} = {font=\scriptsize\color{gray}},
    column{6} = {font=\scriptsize\color{gray}},
    column{8} = {font=\scriptsize\color{gray}},
    column{10} = {font=\scriptsize\color{gray}},
    row{1} = {font=\scriptsize\bfseries},
}
\hline[1pt]
model & reasoning effort & \textsc{combat-arms} & $\pm$ & \textsc{combat-medic} & $\pm$ & \textsc{cyber} & $\pm$ & \textsc{mil-bench-5k} & $\pm$ \\
\hline[0.5pt]
GPT-5 & low & 1.18 & 16.06 & 32.77 & 23.68 & 65.19 & 62.03 & 4.14 & 1.78 \\
EdgeRunner 20B & low & -4.25 & 16.76 & 22.46 & 20 & 104.44 & 75.19 & 9.83 & 1.87 \\
\hline[0.5pt]
GPT-5  & medium & 0 & 15.83 & 0 & 19.58 & 0 & 41.38 & 0 & 1.73 \\
EdgeRunner 20B & medium & -26.7 & 14.46 & 32 & 21.95 & 39.26 & 56.33 & -5.41 & 1.78 \\
\hline[0.5pt]
GPT-5 & high & 10.27 & 17.38 & -10.31 & 18.6 & 56.67 & 58.47 & -1.25 & 1.7 \\
EdgeRunner 20B & high & -27.88 & 13.96 & 51.85 & 25.52 & 43.33 & 57.96 & -6.29 & 1.77 \\
\hline[1pt]
\end{tblr}
\end{table*}

\begin{table*}[htbp]
\caption[]{p-values for the EdgeRunner 20B model versus GPT-5. If the p-value is less than 0.05, then a winner is named. Otherwise, the outcome is considered statistically insignificant and is called a tie. The p-values are based on the absolute score, not on the relative metrics presented in Table \ref{tab:mil-results} and Figure \ref{fig:mil-results}. Here a label of \textsc{win} means that EdgeRunner 20B had a statistically significant win, whereas \textsc{loss} means that GPT-5 had a statistically-significant win.}
\vspace{3pt}
\label{tab:mil-p-values}
\centering
\small
\begin{tblr}{
    colspec={ccccccccc},
    row{1}={font=\bfseries},
    rows={m},
}
\hline[1pt]
reasoning effort & \SetCell[c=2]{c} \textsc{combat-arms} & & \SetCell[c=2]{c} \textsc{combat-medic} & & \SetCell[c=2]{c} \textsc{cyber} & & \SetCell[c=2]{c} \textsc{mil-bench-5k} & \\
\hline[0.5pt]
Low & \SetCell{bg=green!10} 0.752 & \SetCell{bg=green!10} \textsc{tie} & \SetCell{bg=green!10} 0.574 & \SetCell{bg=green!10} \textsc{tie} & \SetCell{bg=green!10} 0.512 & \SetCell{bg=green!10} \textsc{tie} & \SetCell{bg=red!10} 0.002 & \SetCell{bg=red!10} \textsc{loss} \\
Medium & \SetCell{bg=green!10} 0.102 & \SetCell{bg=green!10} \textsc{tie} & \SetCell{bg=green!10} 0.082 & \SetCell{bg=green!10} \textsc{tie} & \SetCell{bg=green!10} 0.420 & \SetCell{bg=green!10} \textsc{tie} & \SetCell{bg=green!30} 0.003 & \SetCell{bg=green!30} \textsc{win} \\
High & \SetCell{bg=green!30} 0.023 & \SetCell{bg=green!30} \textsc{win} & \SetCell{bg=red!10} 0.002 & \SetCell{bg=red!10} \textsc{loss} & \SetCell{bg=green!10} 0.805 & \SetCell{bg=green!10} \textsc{tie} & \SetCell{bg=green!30} 0.005 & \SetCell{bg=green!30} \textsc{win} \\
\hline[1pt]
\end{tblr}
\end{table*}

The EdgeRunner 20B fine-tuned model is within statistical significance or is better than GPT-5 in all cases except for \textsc{combat-medic} with high reasoning and \textsc{mil-bench-5k} with low reasoning. In the cases of \textsc{combat-arms} with high reasoning and \textsc{mil-bench-5k} with medium and high reasoning, the EdgeRunner 20B model has statistically-significant lower error than GPT-5.

As reasoning effort increases, one might expect error to always decrease, but this is not always the case. In a few cases, errors increase when increasing the reasoning effort, which has also been observed by other researchers \cite{hassid2025dontoverthinkitpreferring,gema2025inversescalingtesttimecompute}. We find the medium reasoning effort setting to be the best mix between accuracy and speed, though it should always remain user-configurable.

\subsection{General Purpose Tasks}

\begin{figure*}[htbp]
    \centering
    \includegraphics[width=0.9\linewidth]{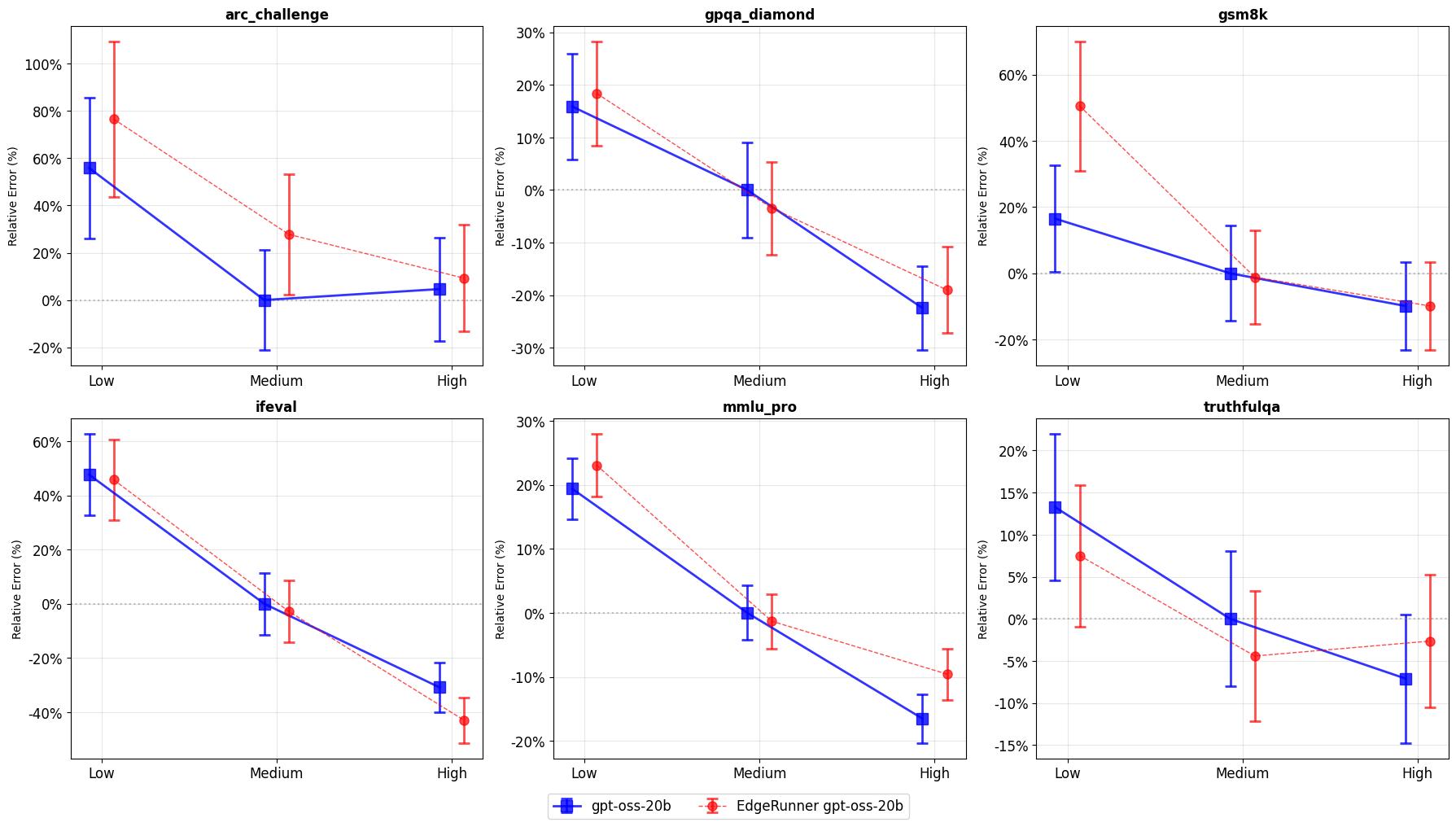}
    \caption{A visual representation of the data presented in Table \ref{tab:gen-results} showing the relative error of EdgeRunner's fine tuned gpt-oss-20b model relative to GPT-5 with medium reasoning. Error bars are given for the standard error. Lower numbers are better (less error).}
    \label{fig:gen-results}
\end{figure*}

When fine tuning, one must prevent regression of the base model's general performance while increasing performance on the new, specialized set of tasks. Relative task performance versus the base model, gpt-oss-20b, is provided in Tables \ref{tab:gen-results} and \ref{tab:gen-p-values} and Figure \ref{fig:gen-results}. For all tasks and reasoning settings except for low reasoning with GSM8k, there is no statistically-significant regression using a p-value of 0.05 measured on the absolute metric values.

\begin{table*}[htbp]
\caption[]{Relative error versus GPT-5 with medium reasoning effort. Lower numbers are better (less error), and negative numbers mean less error than GPT-5 with medium reasoning. }
\label{tab:gen-results}
\vspace{3pt}
\centering
\small
\begin{tblr}{
    colspec={lX[-1,c]X[-1,c]X[-1,c]X[-1,c]X[-1,c]X[-1,c]X[-1,c]X[-1,c]X[-1,c]X[-1,c]X[-1,c]X[-1,c]X[-1,c]},
    rows={m},
    column{1} = {font=\scriptsize},
    column{2} = {font=\scriptsize},
    column{4} = {font=\scriptsize\color{gray}},
    column{6} = {font=\scriptsize\color{gray}},
    column{8} = {font=\scriptsize\color{gray}},
    column{10} = {font=\scriptsize\color{gray}},
    column{12} = {font=\scriptsize\color{gray}},
    column{14} = {font=\scriptsize\color{gray}},
    row{1} = {font=\tiny\bfseries},
}
\hline[1pt]
model & reasoning effort & arc challenge & $\pm$ & gpqa diamond & $\pm$ & gsm8k & $\pm$ & ifeval & $\pm$ & mmlu pro & $\pm$ & truthfulqa & $\pm$ \\
\hline[0.5pt]
gpt-oss-20b & low & 55.86 & 29.81 & 15.89 & 10.03 & 16.52 & 16.06 & 47.67 & 15.05 & 19.40 & 4.78 & 13.27 & 8.70 \\
EdgeRunner 20B & low & 76.57 & 32.94 & 18.38 & 9.92 & 50.58 & 19.53 & 45.80 & 14.92 & 23.07 & 4.89 & 7.52 & 8.41 \\
\hline[0.5pt]
gpt-oss-20b & medium & 0.00 & 21.19 & 0.00 & 9.04 & 0.00 & 14.35 & 0.00 & 11.51 & 0.00 & 4.26 & 0.00 & 8.03 \\
EdgeRunner 20B & medium & 27.79 & 25.54 & -3.43 & 8.83 & -1.16 & 14.16 & -2.83 & 11.35 & -1.31 & 4.24 & -4.41 & 7.77 \\
\hline[0.5pt]
gpt-oss-20b & high & 4.63 & 21.88 & -22.43 & 8.01 & -9.86 & 13.34 & -30.84 & 9.16 & -16.51 & 3.80 & -7.09 & 7.64 \\
EdgeRunner 20B & high & 9.26 & 22.57 & -19.00 & 8.20 & -9.86 & 13.34 & -42.97 & 8.38 & -9.57 & 3.99 & -2.64 & 7.87 \\
\hline[1pt]
\end{tblr}
\end{table*}

\begin{table*}[htbp]
\caption[]{p-values for the EdgeRunner 20B model versus gpt-oss-20b across general-purpose tasks. If the p-value is less than 0.05, then a winner is named. Otherwise, the outcome is considered statistically insignificant and is called a tie. The p-values are based on the absolute score, not on the relative metrics presented in Table \ref{tab:gen-results} and Figure \ref{fig:gen-results}. Here a label of \textsc{win} means that EdgeRunner 20B had a statistically significant win, whereas \textsc{loss} means that gpt-oss-20b had a statistically-significant win.}
\label{tab:gen-p-values}
\vspace{3pt}
\centering
\small
\begin{tblr}{
    colspec={ccccccccc},
    row{1}={font=\bfseries},
    rows={m},
}
\hline[1pt]
reasoning effort & \SetCell[c=2]{c} arc challenge &  & \SetCell[c=2]{c} gpqa diamond &  & \SetCell[c=2]{c} gsm8k &  & \SetCell[c=2]{c} ifeval &  & \SetCell[c=2]{c} mmlu pro &  & \SetCell[c=2]{c} truthfulqa &  \\
\hline[0.5pt]
Low & \SetCell{bg=green!10} 0.4428 & \SetCell{bg=green!10} \textsc{tie} & \SetCell{bg=green!10} 0.7892 & \SetCell{bg=green!10} \textsc{tie} & \SetCell{bg=red!10} 0.0369 & \SetCell{bg=red!10} \textsc{loss} & \SetCell{bg=green!10} 0.8838 & \SetCell{bg=green!10} \textsc{tie} & \SetCell{bg=green!10} 0.4118 & \SetCell{bg=green!10} \textsc{tie} & \SetCell{bg=green!10} 0.4850 & \SetCell{bg=green!10} \textsc{tie} \\
Medium & \SetCell{bg=green!10} 0.2184 & \SetCell{bg=green!10} \textsc{tie} & \SetCell{bg=green!10} 0.7027 & \SetCell{bg=green!10} \textsc{tie} & \SetCell{bg=green!10} 0.9351 & \SetCell{bg=green!10} \textsc{tie} & \SetCell{bg=green!10} 0.8057 & \SetCell{bg=green!10} \textsc{tie} & \SetCell{bg=green!10} 0.7585 & \SetCell{bg=green!10} \textsc{tie} & \SetCell{bg=green!10} 0.5791 & \SetCell{bg=green!10} \textsc{tie} \\
High & \SetCell{bg=green!10} 0.8315 & \SetCell{bg=green!10} \textsc{tie} & \SetCell{bg=green!10} 0.7016 & \SetCell{bg=green!10} \textsc{tie} & \SetCell{bg=green!10} 1.0000 & \SetCell{bg=green!10} \textsc{tie} & \SetCell{bg=green!10} 0.2272 & \SetCell{bg=green!10} \textsc{tie} & \SetCell{bg=green!10} 0.0885 & \SetCell{bg=green!10} \textsc{tie} & \SetCell{bg=green!10} 0.5722 & \SetCell{bg=green!10} \textsc{tie} \\
\hline[1pt]
\end{tblr}
\end{table*}

\subsection{Hyperparameter Analysis}

\begin{table*}[htbp]
\caption[]{Nine fine-tuning runs on gpt-oss-20b with varying hyperparameters, including the reasoning inference setting (``reason''), whether synthetically-generated reasoning traces were included in the training data (``think''), the learning rate (``LR''), the global batch size (``GBS''), the number of training epochs (``epochs''), and the chat template used (``templ''). Results are given as the error relative to the medium reasoning setting of the EdgeRunner 20B model, called ``ER20b'' here. The top section of results compares the EdgeRunner 20B model, which used the alpaca chat template, to a run with a similar number of updates but that used the gpt-oss-20b template. The second section of results compares the use of synthetic reasoning traces to their exclusion. The third section compares a few learning rates, both with and without synthetic reasoning traces.}
\label{tab:ablations}
\vspace{3pt}
\centering
\scriptsize
\begin{tblr}{
    colspec={cccccccccccccc},
    rows={m},
    row{1} = {font=\tiny\bfseries},
}
\hline[1pt]
\# & reason & think & LR & GBS & epochs & templ & combat-arms & $\pm$ & combat-medic & $\pm$ & cyber & $\pm$ & mil-bench-5k & $\pm$ \\
\hline[0.5pt]
ER20b & low & no & 1e-6 & 1536 & 10 & alp & \SetCell{bg=red!10} 30.63 & \SetCell{bg=red!10} 27.57 & \SetCell{bg=green!20} -7.23 & \SetCell{bg=green!20} 11.73 & \SetCell{bg=red!60} 46.81 & \SetCell{bg=red!60} 52.45 & \SetCell{bg=red!20} 16.11 & \SetCell{bg=red!20} 2.15 \\
ER20b & medium & no & 1e-6 & 1536 & 10 & alp & 0.00 & 22.98 & 0.00 & 13.02 & 0.00 & 39.49 & 0.00 & 2.01 \\
ER20b & high & no & 1e-6 & 1536 & 10 & alp & \SetCell{bg=green!50} -1.61 & \SetCell{bg=green!50} 22.29 & \SetCell{bg=red!30} 15.03 & \SetCell{bg=red!30} 15.24 & \SetCell{bg=red!10} 2.93 & \SetCell{bg=red!10} 40.63 & \SetCell{bg=green!50} -0.92 & \SetCell{bg=green!50} 2.00 \\
1 & low & no & 1e-6 & 1024 & 8 & gpt & \SetCell{bg=red!20} 43.84 & \SetCell{bg=red!20} 29.09 & \SetCell{bg=red!10} 6.53 & \SetCell{bg=red!10} 12.76 & \SetCell{bg=red!90} 74.73 & \SetCell{bg=red!90} 64.22 & \SetCell{bg=red!20} 18.09 & \SetCell{bg=red!20} 2.17 \\
1 & medium & no & 1e-6 & 1024 & 8 & gpt & \SetCell{bg=red!10} 14.11 & \SetCell{bg=red!10} 24.60 & \SetCell{bg=green!90} -24.83 & \SetCell{bg=green!90} 10.35 & \SetCell{bg=red!50} 37.23 & \SetCell{bg=red!50} 51.79 & \SetCell{bg=red!10} 3.30 & \SetCell{bg=red!10} 2.03 \\
1 & high & no & 1e-6 & 1024 & 8 & gpt & \SetCell{bg=red!10} 26.52 & \SetCell{bg=red!10} 26.77 & \SetCell{bg=green!50} -13.75 & \SetCell{bg=green!50} 12.25 & \SetCell{bg=red!20} 12.50 & \SetCell{bg=red!20} 47.31 & \SetCell{bg=red!10} 2.98 & \SetCell{bg=red!10} 2.04 \\
\hline[0.5pt]
2 & low & no & 1e-6 & 1024 & 1 & gpt & \SetCell{bg=red!20} 38.93 & \SetCell{bg=red!20} 28.73 & \SetCell{bg=red!30} 15.03 & \SetCell{bg=red!30} 13.88 & \SetCell{bg=red!50} 40.43 & \SetCell{bg=red!50} 50.90 & \SetCell{bg=red!20} 18.25 & \SetCell{bg=red!20} 2.17 \\
2 & medium & no & 1e-6 & 1024 & 1 & gpt & \SetCell{bg=red!10} 7.50 & \SetCell{bg=red!10} 23.74 & \SetCell{bg=green!20} -5.94 & \SetCell{bg=green!20} 12.47 & \SetCell{bg=green!50} -6.38 & \SetCell{bg=green!50} 37.67 & \SetCell{bg=red!10} 1.56 & \SetCell{bg=red!10} 2.01 \\
2 & high & no & 1e-6 & 1024 & 1 & gpt & \SetCell{bg=red!10} 21.52 & \SetCell{bg=red!10} 25.40 & \SetCell{bg=green!40} -11.77 & \SetCell{bg=green!40} 11.93 & \SetCell{bg=red!30} 21.81 & \SetCell{bg=red!30} 46.64 & \SetCell{bg=red!10} 3.72 & \SetCell{bg=red!10} 2.05 \\
3 & low & yes & 1e-6 & 1024 & 1 & gpt & \SetCell{bg=red!20} 43.84 & \SetCell{bg=red!20} 29.52 & \SetCell{bg=red!20} 11.07 & \SetCell{bg=red!20} 13.61 & \SetCell{bg=red!60} 49.73 & \SetCell{bg=red!60} 56.17 & \SetCell{bg=red!20} 22.57 & \SetCell{bg=red!20} 2.22 \\
3 & medium & yes & 1e-6 & 1024 & 1 & gpt & \SetCell{bg=red!10} 14.11 & \SetCell{bg=red!10} 24.77 & \SetCell{bg=green!40} -11.77 & \SetCell{bg=green!40} 12.28 & \SetCell{bg=red!30} 21.81 & \SetCell{bg=red!30} 47.38 & \SetCell{bg=red!10} 7.75 & \SetCell{bg=red!10} 2.08 \\
3 & high & yes & 1e-6 & 1024 & 1 & gpt & \SetCell{bg=red!10} 33.13 & \SetCell{bg=red!10} 27.94 & \SetCell{bg=green!10} -5.24 & \SetCell{bg=green!10} 13.11 & \SetCell{bg=red!30} 18.62 & \SetCell{bg=red!30} 44.91 & \SetCell{bg=red!10} 7.25 & \SetCell{bg=red!10} 2.09 \\
\hline[0.5pt]
5 & low & yes & 1e-6 & 256 & 1 & gpt & \SetCell{bg=red!20} 56.25 & \SetCell{bg=red!20} 31.30 & \SetCell{bg=red!90} 33.92 & \SetCell{bg=red!90} 15.53 & \SetCell{bg=red!60} 43.62 & \SetCell{bg=red!60} 50.60 & \SetCell{bg=red!20} 26.69 & \SetCell{bg=red!20} 2.25 \\
5 & medium & yes & 1e-6 & 256 & 1 & gpt & \SetCell{bg=red!10} 34.73 & \SetCell{bg=red!10} 27.21 & \SetCell{bg=green!10} -4.66 & \SetCell{bg=green!10} 12.14 & \SetCell{bg=red!20} 9.31 & \SetCell{bg=red!20} 41.73 & \SetCell{bg=red!20} 13.37 & \SetCell{bg=red!20} 2.12 \\
5 & high & yes & 1e-6 & 256 & 1 & gpt & \SetCell{bg=red!20} 45.54 & \SetCell{bg=red!20} 29.42 & \SetCell{bg=red!50} 20.16 & \SetCell{bg=red!50} 14.85 & \SetCell{bg=red!20} 9.31 & \SetCell{bg=red!20} 41.01 & \SetCell{bg=red!20} 15.19 & \SetCell{bg=red!20} 2.16 \\
6 & medium & no & 5e-6 & 256 & 1 & gpt & \SetCell{bg=red!20} 63.66 & \SetCell{bg=red!20} 33.17 & -  & -  & -  & - & \SetCell{bg=red!30} 30.99 & \SetCell{bg=red!30} 2.32 \\
7 & medium & yes & 5e-6 & 256 & 1 & gpt & \SetCell{bg=red!30} 162.86 & \SetCell{bg=red!30} 47.87 & - & - & - & - & \SetCell{bg=red!40} 56.09 & \SetCell{bg=red!40} 2.53 \\
8 & medium & yes & 1e-5 & 256 & 1 & gpt & \SetCell{bg=red!90} 515.09 & \SetCell{bg=red!90} 102.28 & - & - & - & - & \SetCell{bg=red!90} 127.98 & \SetCell{bg=red!90} 3.36 \\
\hline[1pt]
\end{tblr}
\end{table*}

We conducted over 40 full fine-tuning runs. Our observations are summarized below, and representative run results are provided in Table \ref{tab:ablations}.

\begin{enumerate}
    \item \textbf{Learning Rate}: We find a learning rate of $1 \times 10^{-6}$ or lower to be best.
    \item \textbf{Global Batch Size}: Large batch sizes performed best, with the highest performance being achieved with between 1,024 and 1,536 records packed to 4,096 sequence length, corresponding to 4.2M to 6.3M tokens per batch. Low batch sizes disturbed the model's reasoning behavior based on our manual analyses.
    \item \textbf{Epochs}: 5-10 epochs is the sweet spot to maximize in-domain task performance while minimizing regression on general tasks, though we conducted a majority of our experiments with 1 epoch to save compute.
    \item \textbf{Chat Template}: Most of our experimentation was performed using the gpt-oss-20b chat template. However, we also tried using a substantially different chat template for training, the alpaca template (see Appendix \ref{sect:app-training-setup}). Inference was still conducted using the original gpt-oss-20b chat template. We found that using the alpaca chat template for training provided a small but significant boost in performance, again due to less degradation of the reasoning behavior of the model.
    \item \textbf{Synthetic Reasoning Traces}: For one series of experiments, we used gpt-oss-120b to generate synthetic reasoning data for the full training dataset, and we conducted some training runs using these reasoning traces. In all cases, we saw degradation versus otherwise-identical runs without the synthetic reasoning data. The objective presented in this work is one of knowledge injection, rather than reasoning improvements, which is why we hypothesize that the simplest data patterns proved most effective.
\end{enumerate}

\subsection{Cloud Cost Analysis}
\label{sect:cost}

Locally hosted, edge models such as EdgeRunner 20B do not incur additional costs each time they are used. By leveraging the user's existing machine, there is not an incremental cost per LLM call. This is an additional key advantage of edge models besides their security, redundancy and privacy. Therefore, the focus in this subsection is solely on the usage costs of cloud-hosted models. We consider four scenarios, and we estimate the cloud costs for each scenario.

\begin{table*}[htbp]
\caption[]{Annual costs for five cloud-based modeling scenarios. Edge models, in contrast, do not incur any of these incremental usage costs after the initial hardware is procured. Details for each scenario are given in section \ref{sect:cost}.}
\label{tab:cost}
\vspace{3pt}
\centering
\footnotesize
\begin{tblr}{
    width=0.9\textwidth,
    colspec={X[2,c]X[2,c]X[1,c]X[1,c]X[1,c]X[1,c]},
    rows={m},
    row{1}={font=\bfseries}
}
\hline[1pt]
Use Case & Assumptions & Avg Input Tok & Avg Output Tok & User Inputs per Workday & Annual Cost per User \\
\hline[1pt]
Web interface flat rate (message/token limits) & \$20/month & --- & --- & --- & \$240 \\
Web interface flat rate (no limits) & \$200/month & --- & --- & --- & \$2,400 \\
Token-based Q\&A & 2k RAG, 1k system prompt & 10,890 & 2,625 & 25 & \$249 \\
Agentic & 15 LLM calls per user input & 627,660 & 21,910 & 25 & \$6,273 \\
Proactive monitoring & runs every minute, 24/7 & 10,000 & 1,500 & --- & \$14,454 \\
\hline[1pt]
\end{tblr}
\end{table*}

\begin{enumerate}
    \item \textbf{Chatbot Subscription}, in which the user pays either (a) a flat rate of \$20 per month with usage limitations or (b) a flat rate of \$200 per month without usage limitations.
    \item \textbf{Token-based Q\&A}, in which the cloud API is used directly and cost is incurred per token. As of July 2025, ChatGPT received 18B weekly messages from 750M weekly active users \cite{Chatterji2025HowPeople,reuters_openai_2025}, so we assume 25 user inputs per workday and 250 workdays per year. We also assume an average conversation length of 5 turns. Using statistics from our four military test sets, we assume an average of 2,625 output tokens including reasoning, as well as an average input of 10,890, which is composed of 2,000 RAG tokens, 1,000 system prompt tokens, and the 7,890 tokens from the preceding conversational context (without reasoning).
    \item \textbf{Agentic and Deep Research}, in which the user tasks an agentic system to perform deep research for authoring an intelligence briefing, to summarize and synthesize military plans and doctrine, to browse the web agentically, to query databases or structured data, to write  and troubleshoot code, or to perform other agentic tasks \cite{wang2025aiagenticprogrammingsurvey}. Using the results for GPT-5 with medium reasoning associated with the MCPMark benchmark \cite{wu2025mcpmarkbenchmarkstresstestingrealistic}, we assume each agentic task to require a total of 627k input tokens and 21.9k output tokens spread across an average of 15 LLM calls. We again assume the user tasks the system 25 times per workday.
    \item \textbf{Proactive and Background Processing}, in which the system runs in a recurring or event-based manner, such as for monitoring signal outputs from upstream systems, watching for anomalous behavior in network traffic, or searching for notable intelligence in social media feeds. Here we assume 10k input tokens and 1.5k output tokens with the model running every minute, 24/7.
\end{enumerate}

For all token-based use cases, we assume \$1.25 per million input tokens and \$10 per million output tokens.

For traditional Q\&A style use cases, the cloud cost per user is tractable, particularly if token limitations are tolerated. However, scaled across millions of Department of War (DoW) members, these costs can become substantial. Moreover, the trend toward agentic and proactive use cases is clear, for which the costs are substantially higher. One can easily imagine a future in which every DoW user employs a swarm of agents that are always working on his or her behalf and that surface information or request user decisions proactively. This inevitable future will be achieved by keeping compute costs low and constant, including with edge compute.

\subsection{Throughput Analysis}

In Table \ref{tab:throughput} we provide token throughput values based on Artificial Analysis \cite{ArtificialAnalysis} for the GPT-5 API and our own benchmarking of gpt-oss-20b using the \texttt{llama-bench} feature in \texttt{llama.cpp} \cite{ggerganovllama.cpp}. For GPT-5 we use the 100-token input setting for token generation speed, which is most similar to the token generation setup from \texttt{llama.cpp}. With a discrete GPU like the Nvidia RTX 5090, gpt-oss-20b token generation speeds can exceed those of the cloud API for GPT-5. On consumer laptops like the MacBook Pro and Air, token generation is slower than that of GPT-5 with the API, but it remains fast enough for most use cases.

\begin{table}[htbp]
\caption[]{Throughput analysis, including prompt processing speed and token generation speed, for GPT-5, taken from Artificial Analysis, and gpt-oss-20b, measured by us.}
\label{tab:throughput}
\vspace{3pt}
\centering
\small
\begin{tblr}{
    colspec={X[1,c]X[1,c]X[1,c]X[1,c]},
    rows={m},
    row{1}={font=\bfseries}
}
\hline[1pt]
Model & Hardware & Prompt tok/s & Gen tok/s \\
\hline[0.5pt]
GPT-5 & API & & 155 \\
\hline[0.5pt]
gpt-oss-20b & Nvidia RTX 5090 & 10,944 & 262 \\
gpt-oss-20b & Macbook Pro M4 Max & 1,659 & 95 \\
gpt-oss-20b & Macbook Air M3 & 365 & 23 \\
\hline[1pt]
\end{tblr}
\end{table}

\section{Related Work}

GPT-5, gpt-oss-20b, and other frontier models are continuously being evaluated for their performance and applicability in real-world domain-specific setups, such as Medicine \citep{wang2025survey, lee2023benefits}, Law \citep{chen2024a} and Science \citep{zhang2024comprehensive}. For this purpose, continuous development and refinement of evaluation benchmarks is necessary for reflecting accurately the capabilities of such models in the domain-specific settings \citep{pmlr-v174-pal22a, chen2021evaluating, andrews2025arescalingagentenvironments}. Fine tuning remains a popular approach, and models like the GPT series from the OpenAI API and the Nova models \cite{agi2025amazonnovafamilymodels} on AWS Bedrock come with API-based fine-tuning capabilities, while open-weight models are often more deeply customized using model training libraries.

Due to the restrictive nature of defense-related studies and applications, publicly-available works that report thorough findings and insights from the investigation of state-of-the-art language models, such as gpt-oss-20b, on military tasks are limited. One notable work is the TRACLM project \citep{ruiz2024fine}, in which the authors proposed a suite of benchmark tasks (\textit{MilBench/MilGLUE} \citep{milglue}) adapted to the military domain and trained a military-focused model based on unclassified relevant documents. Besides this work, the GovBench team created the JointStaffBench dataset \cite{GovBench:JointStaffBench} for evaluating LLMs on joint publication knowledge, though the full dataset is closed to non-military personnel.

\section{Conclusion and Next Steps}

In this work we've presented Edgerunner 20B, a highly capable, military-focused, air-gapped model that can run on edge devices. We've shown its relative performance on military-specific and general-purpose benchmarks, and we've shown its competitiveness against the frontier proprietary model GPT-5. Throughout our analysis we include the approaches we used to build our datasets, benchmarks and models, as well as provide additional reports regarding the cost and hardware efficiency dimensions.

We continue our path towards covering all gaps and improving security and performance. On the evaluation front, we are currently working on a host of additional test sets, including gold test sets for combat medics, tactical combat arms, operational combat arms, vehicle maintenance, and aircraft maintenance. Over time we will develop a robust suite of evaluations and an associated leaderboard.

Additionally, we are currently preparing a series of tests and metrics for measuring the rates of request deflection and resolution, as well as novel approaches for improving model performance based on these criteria. Refusals stem from user requests that can be interpreted by the models as harmful or generally ill-intentioned, which could hold true for every-day consumer tasks. However, defense deployments and mission-critical use-cases require significantly different treatments. We must ensure that our models are always reliable in real-world military use-cases without sacrificing security and while adhering to applicable laws of armed conflict and rules of engagement.

Lastly, we are studying a variety of models that can be used for defense purposes and are suitable for edge devices, including various model sizes that can be as small as 0.5-1B parameters. To this end, we are also currently experimenting with pre-training approaches utilizing our internal large-scale military corpus that we have been building in our labs, and we are investigating post-training techniques (e.g. Reinforcement Learning) for enhancing performance in complex real-world deployments, with the goal of developing state-of-the-art models that excel in domain-specific and general-purpose benchmarks, all while living in air-gapped edge devices.

\bibliography{bibliography}

\begin{thebibliography}{39}
\providecommand{\natexlab}[1]{#1}
\providecommand{\url}[1]{\texttt{#1}}
\expandafter\ifx\csname urlstyle\endcsname\relax
  \providecommand{\doi}[1]{doi: #1}\else
  \providecommand{\doi}{doi: \begingroup \urlstyle{rm}\Url}\fi

\bibitem[gpt(2025)]{gpt5-system-card}
Gpt-5 system card.
\newblock Technical report, OpenAI, 2025.
\newblock URL \url{https://cdn.openai.com/gpt-5-system-card.pdf}.

\bibitem[AGI et~al.(2025)AGI, Langford, Shah, Gupta, Bhatter, Goyal, Mathur, Mohanty, Kumar, Sethi, Komma, Pena, Jain, Kunysz, Opyrchal, Singh, Rawal, Prasad, de~Gispert, Kumar, Aryamane, Nair, M, Iyengar, Shanbhogue, He, Cervone, Loeb, Zhang, Fu, Lisnichenko, Zhipa, Potamianos, Kebarighotbi, Daronkolaei, Parmesh, Samra, Khan, Rez, Saffari, Agarwalla, Jhindal, Mamidala, Asmro, Ballakur, Mishra, Sridharan, Dubinina, Lenz, Doerr, Keating, Leaver, Smith, Wirth, Davey, Rosenbaum, Sohn, Chan, Chakrabarti, Ramakrishna, Roy, Iyer, Narayan-Chen, Yennu, Dabrowska, Gawlowska, Rumshisky, Turek, Deoras, Bezruchkin, Prasad, Dewan, Kiran, Gupta, Galstyan, Manoharan, Biswas, Mandal, Gupta, Pathan, Nagarajan, Rajasekaram, Sundararajan, Ganesan, Swaminathan, Mouchtaris, Champeau, Ray, Jaiswal, Sharma, Keefer, Muthiah, Leon-Millan, Koopman, Li, Biggs, Ott, Vinzamuri, Venkatesh, Ganesh, Vasani, Byrne, Hsu, Wang, King, Gorny, Feng, Zheng, Paul, Sun, Luo, Chen, Xie, Yu, Jugan, Panosh, Collins, Thompson, Karakus, Liu, Lambrecht,
  Lin, Wang, Yuan, Loyda, Walczak, Choppa, Prakash, Meas, Peris, Recaido, Xu, Sharma, Kernan, Thanapirom, Su, Xu, Yin, Ye, Tao, Parameshwara, Chang, Li, Hench, Tran, Dupuy, Davis, DiPersio, Christodoulopoulos, Li, Chen, Bovi, Chung, Hawkins, Harris, Ropell, He, Joo, Hwang, Rosen, Elkind, Pressel, Zhang, Kimball, Sorokin, Goodell, Modolo, Zhu, Suresh, Ragha, Filimonov, Kune, Rodriguez, Hazarika, Ram, Parkar, Patel, Desai, Rajput, Sule, Singh, Genzel, Goldenberg, He, Hanciu, Tharmal, Siankovich, Cikovic, Abraham, Sabir, Olson, Steven, Barut, Jackson, Wu, Chen, Mahalingam, Triefenbach, Yang, Liu, Wu, Tavakoli, Khozeimeh, Niu, Hieber, Li, Elbey, Krebs, Saupe, Sprünken, Fan, Khan, Vincenzo, Kang, Ding, He, Yeung, Qaddoumi, Karamanolakis, Huybrechts, Maddali, Iglesias, McShane, Sahin, Huang, Kwon, Sigurdsson, Chadha, Kosuru, Fuerstenau, Hah, Maideen, Hosokawa, Liu, Hsu, Wang, Li, Yang, Zhu, Fan, Singh, Kaluvala, Saeed, Xie, Feng, Luo, Pei, Nielsen, Ilati, Patel, Li, Lin, Raza, Cullinan, Kiss, Thangamani, Fadnavis,
  Sorodoc, Ertuerk, Yemialyanava, Soni, Jelal, Tse, FitzGerald, Zhao, Rothgeb, Lee, Jung, Debski, Tomczak, Jeun, Sanders, Crowley, Lee, Paidy, Tiwari, Farmer, Solinsky, Lau, Savareese, Zagorski, Dai, Jiacheng, Gu, Li, Jian, Zheng, Lu, Wang, Dai, Mo, Xu, Liang, Yang, Logan, Majmudar, Liu, Miao, Yi, Jin, Kao, Wang, Wang, Pemberton, Carlson, Blundell, Chin-Jew, He, Ho, Hueser, Lunt, Lee, Tan, Chatterjee, Gaspers, Wang, Fang, Tang, Wan, Wu, Wang, Shi, Chiu, Satriano, Yee, Dhamala, Bansal, Zhen, Chang, Lin, Raman, Sathyendra, Moroe, Bhandarkar, Kothari, Owczarzak, Gopalswamy, Ravi, Ramakrishnan, Arumugam, Mehta, Konczalska, Ravikumar, Tran, Qin, Li, Li, Kulkarni, Rodrigues, Patel, Abboud, Hajebi, Reiter, Schultz, Anisetty, Kotnana, Li, Channamallikarjuna, Jakubczyk, Pierewoj, Pal, Srivastav, Bannerman, Poddar, Prasad, Tseng, Naik, Vankadara, Minorics, Liu, Lausen, Ribeiro, Zhang, Gehorsam, Qi, Bauer, Knapp, Zeng, Tong, Wong, Chen, Rudnicki, Namazifar, Jaliminche, Tanke, Gupta, Ahlawat, Khanuja, Sundaram, Leyk,
  Momotko, Boese, Dreyer, Mueller, Fu, Górski, Mastalerczyk, Mora, Johnson, Scott, Wen, Barysau, Boumerdassi, Krishnan, Gupta, Hirani, Kulkarni, Narayanasamy, Bradford, Gens, Burke, Jin, Chen, Denkowski, Heymel, Krestyaninov, Obirek, Wichorowska, Miotk, Watroba, Hong, Yu, Liu, Gouda, El-Shabani, Ghavamzadeh, Bansal, Ziyadi, Xia, Susanj, Bhasin, Goswami, Belgamwar, Anastassacos, Bergeron, Jain, Jain, Chopparapu, Xu, Strom, Malandrakis, Mishra, Parkhi, Mehrabi, Sant, Gupta, Sekhar, Rajeev, Chidambaram, Dhar, Bhagwagar, Konforty, Babu, Razavi, Majumder, Dar, Hsu, Kvitca, Pandey, Seegmiller, Lange, Ferraro, Motwani, Kharazmi, Wang, Liu, Bradtke, Götz, Zhou, Wang, Poskart, Sonawane, Natarajan, Ramadorai, Shah, Nirantar, Chavali, Wanigasekara, Saraf, Dey, Pant, Pradhan, Patel, Dadlani, Sadha, Dong, Hu, Qiaozi, Gao, Liu, Lam, Do, Manmatha, Willis, Liu, Ellert, Kalinski, Attrach, Prasad, Prasad, Kunani, Gupta, Sharma, Tewari, Baskaran, Singh, Gupta, Reddy, Das, Chada, Mahesh, Chandrasekaran, Nallapati, Xue,
  Gangadharaiah, Rachakonda, Zhang, Blloshmi, Agrawal, Enyedi, Lowe, Shrestha, Piramuthu, Asad, Khanna, Mukherjee, Mittal, Prasad, Kumar, Diamant, Gupta, Li, Li, Fegade, Zhang, Arbow, Chen, Gabbard, Hoium, King, Iyer, Malick, Movaghati, Balakavi, Jakka, Paruvelli, Jayanthi, Mujumdar, Kapoor, Beygi, Dingliwal, Soltan, Ricklin, Tucker, Sinha, Choudhary, Tan, Broscheit, Schulter, Agarwal, Atluri, Valstar, Shankar, Sanyukta, Khanna, Khetrapal, Janakiraman, Shah, Akolkar, Giri, Khandelwal, Pawar, Sahu, Huang, Ra, Gopal, Dobroshinsky, Saba, Roy, Lal, Ananthakrishnan, Li, Srijan, Bhide, Tang, Zha, Oraby, Mostafa, Li, Bharathi, Prakash, Huang, Yembarwar, Pansare, Subramanian, Joshi, Liu, Tang, Chandak, Garg, Katiyar, Mehta, Srivastav, Yang, S, Choudhary, Senger, Babb, Moeini, Deng, Loganathan, Domagala, Narkar, Wadhwa, Zhang, Jiang, Trenous, Sarkar, Saha, Reddy, Dokania, Sandiri, Matsoukas, Bodapati, Wdaru, Venkateshdatta, Ronanki, Veeravanallur, Venkatapathy, Sankaraguru, Gorantla, Karuturi, Schroedl, Rongali,
  Kundu, Shakiah, Tiwari, Bharti, Sami, Mathew, Yu, Kim, Malode, Riel, Palod, Roy, Furqhan, Chung, Yoshitani, Yang, Chillakura, Bajwa, Lajumoke, Tran, Gueudre, Jung, Li, Seemman, Leffel, Xiang, Patel, Domhan, Falke, Guo, Li, Horszczaruk, Jedynak, Kulkarni, Marin, Metrycki, Wang, Jain, Singh, Chirimar, Gupta, Shah, Deshpande, Gunjal, Srikeshava, Vivek, Bharadwaj, Gangal, Kumar, Elango, Ordonez, Soto, Radhakrishnan, Patel, Singh, Kolanuvada, Kumar, Auvray, Cartillier, Ponzo, Peng, Khandelwal, Naik, Sahasrabudhe, Korolev, Gokuladas, Madan, Subramanian, Cevher, Gupta, Hamza, Zhang, Ruan, Cheng, Zhang, Zhao, Yao, Ouyang, Dashner, Campbell, Lin, Martin, Pearson, Jiang, Lu, Shi, Peng, Gao, Jiang, Fei, Wang, Zhou, Feng, Zhao, Wang, Li, Zhang, Wang, Fu, Yuan, Wang, Rao, Tavizon, Rossiytsev, Chen, Liu, Zou, Park, Versley, Zhang, Patel, Lu, Pan, Yi-Hsiang, Lai, Hu, Wang, Zhou, Xiang, Shi, Wang, Galatzer, Wang, Shen, Sun, Purwatama, Yue, Wu, Gu, Wang, Zeng, Chen, Zhou, Xie, Guy, Ambrozinski, Cai, Zhang, Wang, Jin, Zhao,
  Li, Luo, Zhang, Fang, Bu, Wang, Li, Wang, Zimeng, Qiu, and Li]{agi2025amazonnovafamilymodels}
AGI, A., Langford, A., Shah, A., Gupta, A., Bhatter, A., Goyal, A., Mathur, A., Mohanty, A., Kumar, A., Sethi, A., Komma, A., Pena, A., Jain, A., Kunysz, A., Opyrchal, A., Singh, A., Rawal, A., Prasad, A. A.~B., de~Gispert, A., Kumar, A., Aryamane, A., Nair, A., M, A., Iyengar, A., Shanbhogue, A. V.~K., He, A., Cervone, A., Loeb, A., Zhang, A., Fu, A., Lisnichenko, A., Zhipa, A., Potamianos, A., Kebarighotbi, A., Daronkolaei, A., Parmesh, A., Samra, A.~K., Khan, A., Rez, A., Saffari, A., Agarwalla, A., Jhindal, A., Mamidala, A., Asmro, A., Ballakur, A., Mishra, A., Sridharan, A., Dubinina, A., Lenz, A., Doerr, A., Keating, A., Leaver, A., Smith, A., Wirth, A., Davey, A., Rosenbaum, A., Sohn, A., Chan, A., Chakrabarti, A., Ramakrishna, A., Roy, A., Iyer, A., Narayan-Chen, A., Yennu, A., Dabrowska, A., Gawlowska, A., Rumshisky, A., Turek, A., Deoras, A., Bezruchkin, A., Prasad, A., Dewan, A., Kiran, A., Gupta, A., Galstyan, A., Manoharan, A., Biswas, A., Mandal, A., Gupta, A., Pathan, A., Nagarajan, A.,
  Rajasekaram, A., Sundararajan, A., Ganesan, A., Swaminathan, A., Mouchtaris, A., Champeau, A., Ray, A., Jaiswal, A., Sharma, A., Keefer, B., Muthiah, B., Leon-Millan, B., Koopman, B., Li, B., Biggs, B., Ott, B., Vinzamuri, B., Venkatesh, B., Ganesh, B., Vasani, B., Byrne, B., Hsu, B., Wang, B., King, B., Gorny, B., Feng, B., Zheng, B., Paul, B., Sun, B., Luo, B., Chen, B., Xie, B., Yu, B., Jugan, B., Panosh, B., Collins, B., Thompson, B., Karakus, C., Liu, C., Lambrecht, C., Lin, C., Wang, C., Yuan, C., Loyda, C., Walczak, C., Choppa, C., Prakash, C.~S., Meas, C.~R., Peris, C., Recaido, C., Xu, C., Sharma, C., Kernan, C., Thanapirom, C., Su, C., Xu, C., Yin, C., Ye, C., Tao, C., Parameshwara, C., Chang, C.-Y., Li, C., Hench, C., Tran, C., Dupuy, C., Davis, C., DiPersio, C., Christodoulopoulos, C., Li, C., Chen, C., Bovi, C.~D., Chung, C., Hawkins, C., Harris, C., Ropell, C., He, C., Joo, D., Hwang, D.~Y., Rosen, D., Elkind, D., Pressel, D., Zhang, D., Kimball, D., Sorokin, D., Goodell, D., Modolo, D., Zhu,
  D., Suresh, D., Ragha, D., Filimonov, D., Kune, D.~F., Rodriguez, D.~R., Hazarika, D., Ram, D., Parkar, D., Patel, D., Desai, D., Rajput, D.~S., Sule, D., Singh, D., Genzel, D., Goldenberg, D., He, D., Hanciu, D., Tharmal, D., Siankovich, D., Cikovic, E., Abraham, E., Sabir, E., Olson, E., Steven, E., Barut, E., Jackson, E., Wu, E., Chen, E., Mahalingam, E., Triefenbach, F., Yang, F., Liu, F., Wu, F., Tavakoli, F., Khozeimeh, F., Niu, F., Hieber, F., Li, F., Elbey, F., Krebs, F., Saupe, F., Sprünken, F., Fan, F., Khan, F., Vincenzo, G.~D., Kang, G., Ding, G., He, G., Yeung, G., Qaddoumi, G., Karamanolakis, G., Huybrechts, G., Maddali, G., Iglesias, G., McShane, G., Sahin, G., Huang, G., Kwon, G., Sigurdsson, G.~A., Chadha, G., Kosuru, G., Fuerstenau, H., Hah, H., Maideen, H., Hosokawa, H., Liu, H., Hsu, H.-K., Wang, H., Li, H., Yang, H., Zhu, H., Fan, H., Singh, H., Kaluvala, H., Saeed, H., Xie, H., Feng, H., Luo, H., Pei, H., Nielsen, H., Ilati, H., Patel, H., Li, H., Lin, H., Raza, H., Cullinan, I.,
  Kiss, I., Thangamani, I., Fadnavis, I., Sorodoc, I.~T., Ertuerk, I., Yemialyanava, I., Soni, I., Jelal, I., Tse, I., FitzGerald, J., Zhao, J., Rothgeb, J., Lee, J., Jung, J., Debski, J., Tomczak, J., Jeun, J., Sanders, J., Crowley, J., Lee, J., Paidy, J.~A., Tiwari, J., Farmer, J., Solinsky, J., Lau, J., Savareese, J., Zagorski, J., Dai, J., Jiacheng, Gu, Li, J., Jian, Zheng, Lu, J., Wang, J., Dai, J., Mo, J., Xu, J., Liang, J., Yang, J., Logan, J., Majmudar, J., Liu, J., Miao, J., Yi, J., Jin, J., Kao, J.-Y., Wang, J., Wang, J., Pemberton, J., Carlson, J., Blundell, J., Chin-Jew, J., He, J., Ho, J., Hueser, J., Lunt, J., Lee, J., Tan, J., Chatterjee, J., Gaspers, J., Wang, J., Fang, J., Tang, J., Wan, J., Wu, J., Wang, J., Shi, J., Chiu, J., Satriano, J., Yee, J., Dhamala, J., Bansal, J., Zhen, K., Chang, K.-W., Lin, K., Raman, K., Sathyendra, K.~M., Moroe, K., Bhandarkar, K., Kothari, K., Owczarzak, K., Gopalswamy, K., Ravi, K., Ramakrishnan, K., Arumugam, K., Mehta, K., Konczalska, K., Ravikumar, K.,
  Tran, K., Qin, K., Li, K., Li, K., Kulkarni, K., Rodrigues, K.~A., Patel, K., Abboud, K., Hajebi, K., Reiter, K., Schultz, K., Anisetty, K., Kotnana, K., Li, K., Channamallikarjuna, K., Jakubczyk, K., Pierewoj, K., Pal, K., Srivastav, K., Bannerman, K., Poddar, L., Prasad, L., Tseng, L., Naik, L., Vankadara, L.~C., Minorics, L., Liu, L., Lausen, L., Ribeiro, L. F.~R., Zhang, L., Gehorsam, L., Qi, L., Bauer, L., Knapp, L., Zeng, L., Tong, L., Wong, L., Chen, L., Rudnicki, M., Namazifar, M., Jaliminche, M., Tanke, M.~L., Gupta, M., Ahlawat, M., Khanuja, M., Sundaram, M., Leyk, M., Momotko, M., Boese, M., Dreyer, M., Mueller, M., Fu, M., Górski, M., Mastalerczyk, M., Mora, M., Johnson, M., Scott, M., Wen, M., Barysau, M., Boumerdassi, M., Krishnan, M., Gupta, M., Hirani, M., Kulkarni, M., Narayanasamy, M., Bradford, M., Gens, M., Burke, M., Jin, M., Chen, M., Denkowski, M., Heymel, M., Krestyaninov, M., Obirek, M., Wichorowska, M., Miotk, M., Watroba, M., Hong, M., Yu, M., Liu, M., Gouda, M., El-Shabani, M.,
  Ghavamzadeh, M., Bansal, M., Ziyadi, M., Xia, N., Susanj, N., Bhasin, N., Goswami, N., Belgamwar, N., Anastassacos, N., Bergeron, N., Jain, N., Jain, N., Chopparapu, N., Xu, N., Strom, N., Malandrakis, N., Mishra, N., Parkhi, N., Mehrabi, N., Sant, N., Gupta, N., Sekhar, N., Rajeev, N., Chidambaram, N.~R., Dhar, N., Bhagwagar, N., Konforty, N., Babu, O., Razavi, O., Majumder, O., Dar, O., Hsu, O., Kvitca, P., Pandey, P., Seegmiller, P., Lange, P., Ferraro, P., Motwani, P., Kharazmi, P., Wang, P., Liu, P., Bradtke, P., Götz, P., Zhou, P., Wang, P., Poskart, P., Sonawane, P., Natarajan, P., Ramadorai, P., Shah, P., Nirantar, P., Chavali, P., Wanigasekara, P., Saraf, P., Dey, P., Pant, P., Pradhan, P., Patel, P., Dadlani, P., Sadha, P.~N., Dong, Q., Hu, Q., Qiaozi, Gao, Liu, Q., Lam, Q., Do, Q., Manmatha, R., Willis, R., Liu, R., Ellert, R., Kalinski, R., Attrach, R.~A., Prasad, R., Prasad, R., Kunani, R., Gupta, R., Sharma, R., Tewari, R., Baskaran, R., Singh, R., Gupta, R., Reddy, R., Das, R., Chada, R.,
  Mahesh, R.~V., Chandrasekaran, R., Nallapati, R., Xue, R., Gangadharaiah, R., Rachakonda, R., Zhang, R., Blloshmi, R., Agrawal, R., Enyedi, R., Lowe, R., Shrestha, R., Piramuthu, R., Asad, R., Khanna, R., Mukherjee, R., Mittal, R., Prasad, R., Kumar, R. M.~V., Diamant, R., Gupta, R., Li, R., Li, R., Fegade, R., Zhang, R., Arbow, R., Chen, R., Gabbard, R., Hoium, R., King, R., Iyer, S., Malick, S., Movaghati, S., Balakavi, S., Jakka, S., Paruvelli, S.~K., Jayanthi, S.~M., Mujumdar, S.~S., Kapoor, S., Beygi, S., Dingliwal, S., Soltan, S., Ricklin, S., Tucker, S., Sinha, S., Choudhary, S., Tan, S., Broscheit, S., Schulter, S., Agarwal, S., Atluri, S., Valstar, S., Shankar, S., Sanyukta, S., Khanna, S., Khetrapal, S., Janakiraman, S., Shah, S., Akolkar, S., Giri, S., Khandelwal, S., Pawar, S., Sahu, S., Huang, S., Ra, S., Gopal, S., Dobroshinsky, S., Saba, S., Roy, S., Lal, S., Ananthakrishnan, S., Li, S., Srijan, S., Bhide, S., Tang, S.~L., Zha, S., Oraby, S., Mostafa, S., Li, S., Bharathi, S., Prakash, S.,
  Huang, S., Yembarwar, S., Pansare, S., Subramanian, S., Joshi, S., Liu, S., Tang, S., Chandak, S., Garg, S., Katiyar, S., Mehta, S., Srivastav, S., Yang, S., S, S.~D., Choudhary, S., Senger, S.~S., Babb, S., Moeini, S., Deng, S., Loganathan, S., Domagala, S., Narkar, S., Wadhwa, S., Zhang, S., Jiang, S., Trenous, S., Sarkar, S., Saha, S., Reddy, S., Dokania, S., Sandiri, S., Matsoukas, S., Bodapati, S., Wdaru, S. H.~R., Venkateshdatta, S.~Y., Ronanki, S., Veeravanallur, S.~R., Venkatapathy, S., Sankaraguru, S., Gorantla, S., Karuturi, S., Schroedl, S., Rongali, S., Kundu, S., Shakiah, S., Tiwari, S., Bharti, S., Sami, S., Mathew, S., Yu, S., Kim, S., Malode, S.~B., Riel, S.~C., Palod, S., Roy, S., Furqhan, S., Chung, T., Yoshitani, T., Yang, T., Chillakura, T., Bajwa, T., Lajumoke, T., Tran, T., Gueudre, T., Jung, T., Li, T., Seemman, T., Leffel, T., Xiang, T., Patel, T., Domhan, T., Falke, T., Guo, T., Li, T., Horszczaruk, T., Jedynak, T., Kulkarni, T., Marin, T., Metrycki, T., Wang, T.-Y., Jain, U.,
  Singh, U., Chirimar, U., Gupta, V., Shah, V., Deshpande, V., Gunjal, V., Srikeshava, V., Vivek, V., Bharadwaj, V., Gangal, V., Kumar, V., Elango, V., Ordonez, V., Soto, V., Radhakrishnan, V., Patel, V., Singh, V., Kolanuvada, V.~V., Kumar, V.~B., Auvray, V., Cartillier, V., Ponzo, V., Peng, V., Khandelwal, V., Naik, V., Sahasrabudhe, V., Korolev, V., Gokuladas, V., Madan, V., Subramanian, V., Cevher, V., Gupta, V., Hamza, W., Zhang, W., Ruan, W., Cheng, W., Zhang, W., Zhao, W., Yao, W., Ouyang, W., Dashner, W., Campbell, W., Lin, W., Martin, W., Pearson, W., Jiang, X., Lu, X., Shi, X., Peng, X., Gao, X., Jiang, X., Fei, X., Wang, X., Zhou, X.~J., Feng, X., Zhao, X., Wang, X., Li, X., Zhang, X., Wang, X., Fu, X., Yuan, X., Wang, X., Rao, Y., Tavizon, Y., Rossiytsev, Y., Chen, Y., Liu, Y., Zou, Y., Park, Y., Versley, Y., Zhang, Y., Patel, Y., Lu, Y.-C., Pan, Y., Yi-Hsiang, Lai, Hu, Y., Wang, Y., Zhou, Y., Xiang, Y., Shi, Y., Wang, Y., Galatzer, Y., Wang, Y., Shen, Y., Sun, Y., Purwatama, Y., Yue, Wu, Gu, Y.,
  Wang, Y., Zeng, Y., Chen, Y., Zhou, Y., Xie, Y., Guy, Y., Ambrozinski, Z., Cai, Z., Zhang, Z., Wang, Z., Jin, Z., Zhao, Z., Li, Z., Luo, Z., Zhang, Z., Fang, Z., Bu, Z., Wang, Z., Li, Z., Wang, Z., Zimeng, Qiu, and Li, Z.
\newblock The amazon nova family of models: Technical report and model card, 2025.
\newblock URL \url{https://arxiv.org/abs/2506.12103}.

\bibitem[AI~Security~Institute()]{UK_AI_Security_Institute_Inspect_AI_Framework_2024}
AI~Security~Institute, U.
\newblock Inspect {AI:} {Framework} for {Large} {Language} {Model} {Evaluations}.
\newblock URL \url{https://github.com/UKGovernmentBEIS/inspect_ai}.

\bibitem[Alexandru et~al.(2025)Alexandru, Calvi, Broomfield, Golden, Dai, Leys, Burger, Bartolo, Engeler, Pisupati, Drane, and Park]{alexandru2025atlaseleneminigeneral}
Alexandru, A., Calvi, A., Broomfield, H., Golden, J., Dai, K., Leys, M., Burger, M., Bartolo, M., Engeler, R., Pisupati, S., Drane, T., and Park, Y.~S.
\newblock Atla selene mini: A general purpose evaluation model, 2025.
\newblock URL \url{https://arxiv.org/abs/2501.17195}.

\bibitem[Andrews et~al.(2025)Andrews, Benhalloum, Bertran, Bettini, Budhiraja, Cabral, Do, Froger, Garreau, Gaya, Laurençon, Lecanu, Malkan, Mekala, Ménard, Mialon, Piterbarg, Plekhanov, Rita, Rusakov, Scialom, Vorotilov, Wang, and Yu]{andrews2025arescalingagentenvironments}
Andrews, P., Benhalloum, A., Bertran, G. M.-T., Bettini, M., Budhiraja, A., Cabral, R.~S., Do, V., Froger, R., Garreau, E., Gaya, J.-B., Laurençon, H., Lecanu, M., Malkan, K., Mekala, D., Ménard, P., Mialon, G., Piterbarg, U., Plekhanov, M., Rita, M., Rusakov, A., Scialom, T., Vorotilov, V., Wang, M., and Yu, I.
\newblock Are: Scaling up agent environments and evaluations, 2025.
\newblock URL \url{https://arxiv.org/abs/2509.17158}.

\bibitem[{Artificial Analysis}(2024)]{ArtificialAnalysis}
{Artificial Analysis}.
\newblock {AI Model \& API Providers Analysis}.
\newblock \url{https://artificialanalysis.ai/}, 2024.
\newblock Accessed: 2025-10-28.

\bibitem[{Axolotl maintainers and contributors}(2023)]{axolotl}
{Axolotl maintainers and contributors}.
\newblock Axolotl: Open source llm post-training, 2023.
\newblock URL \url{https://github.com/axolotl-ai-cloud/axolotl}.

\bibitem[Chatterji et~al.(2025)Chatterji, Cunningham, Deming, Hitzig, Ong, Shan, and Wadman]{Chatterji2025HowPeople}
Chatterji, A., Cunningham, T., Deming, D.~J., Hitzig, Z., Ong, C., Shan, C.~Y., and Wadman, K.
\newblock How people use {ChatGPT}.
\newblock Working Paper w34255, National Bureau of Economic Research, Cambridge, MA, 2025.
\newblock URL \url{https://doi.org/10.3386/w34255}.

\bibitem[Chen et~al.(2021)Chen, Tworek, Jun, Yuan, de~Oliveira~Pinto, Kaplan, Edwards, Burda, Joseph, Brockman, Ray, Puri, Krueger, Petrov, Khlaaf, Sastry, Mishkin, Chan, Gray, Ryder, Pavlov, Power, Kaiser, Bavarian, Winter, Tillet, Such, Cummings, Plappert, Chantzis, Barnes, Herbert-Voss, Guss, Nichol, Paino, Tezak, Tang, Babuschkin, Balaji, Jain, Saunders, Hesse, Carr, Leike, Achiam, Misra, Morikawa, Radford, Knight, Brundage, Murati, Mayer, Welinder, McGrew, Amodei, McCandlish, Sutskever, and Zaremba]{chen2021evaluating}
Chen, M., Tworek, J., Jun, H., Yuan, Q., de~Oliveira~Pinto, H.~P., Kaplan, J., Edwards, H., Burda, Y., Joseph, N., Brockman, G., Ray, A., Puri, R., Krueger, G., Petrov, M., Khlaaf, H., Sastry, G., Mishkin, P., Chan, B., Gray, S., Ryder, N., Pavlov, M., Power, A., Kaiser, L., Bavarian, M., Winter, C., Tillet, P., Such, F.~P., Cummings, D., Plappert, M., Chantzis, F., Barnes, E., Herbert-Voss, A., Guss, W.~H., Nichol, A., Paino, A., Tezak, N., Tang, J., Babuschkin, I., Balaji, S., Jain, S., Saunders, W., Hesse, C., Carr, A.~N., Leike, J., Achiam, J., Misra, V., Morikawa, E., Radford, A., Knight, M., Brundage, M., Murati, M., Mayer, K., Welinder, P., McGrew, B., Amodei, D., McCandlish, S., Sutskever, I., and Zaremba, W.
\newblock Evaluating large language models trained on code, 2021.

\bibitem[Chen et~al.(2024)Chen, Ma, Zhang, Hao, Yan, Nourbakhsh, Yang, McAuley, Petzold, and Wang]{chen2024a}
Chen, Z., Ma, J., Zhang, X., Hao, N., Yan, A., Nourbakhsh, A., Yang, X., McAuley, J., Petzold, L.~R., and Wang, W.~Y.
\newblock A survey on large language models for critical societal domains: Finance, healthcare, and law.
\newblock \emph{Transactions on Machine Learning Research}, 2024.
\newblock ISSN 2835-8856.
\newblock URL \url{https://openreview.net/forum?id=upAWnMgpnH}.
\newblock Survey Certification.

\bibitem[Clark et~al.(2018)Clark, Cowhey, Etzioni, Khot, Sabharwal, Schoenick, and Tafjord]{clark2018thinksolvedquestionanswering}
Clark, P., Cowhey, I., Etzioni, O., Khot, T., Sabharwal, A., Schoenick, C., and Tafjord, O.
\newblock Think you have solved question answering? try arc, the ai2 reasoning challenge, 2018.
\newblock URL \url{https://arxiv.org/abs/1803.05457}.

\bibitem[Cobbe et~al.(2021)Cobbe, Kosaraju, Bavarian, Chen, Jun, Kaiser, Plappert, Tworek, Hilton, Nakano, Hesse, and Schulman]{cobbe2021trainingverifierssolvemath}
Cobbe, K., Kosaraju, V., Bavarian, M., Chen, M., Jun, H., Kaiser, L., Plappert, M., Tworek, J., Hilton, J., Nakano, R., Hesse, C., and Schulman, J.
\newblock Training verifiers to solve math word problems, 2021.
\newblock URL \url{https://arxiv.org/abs/2110.14168}.

\bibitem[Daniel~Han \& team(2023)Daniel~Han and team]{unsloth}
Daniel~Han, M.~H. and team, U.
\newblock Unsloth, 2023.
\newblock URL \url{http://github.com/unslothai/unsloth}.

\bibitem[Gema et~al.(2025)Gema, Hägele, Chen, Arditi, Goldman-Wetzler, Fraser-Taliente, Sleight, Petrini, Michael, Alex, Minervini, Chen, Benton, and Perez]{gema2025inversescalingtesttimecompute}
Gema, A.~P., Hägele, A., Chen, R., Arditi, A., Goldman-Wetzler, J., Fraser-Taliente, K., Sleight, H., Petrini, L., Michael, J., Alex, B., Minervini, P., Chen, Y., Benton, J., and Perez, E.
\newblock Inverse scaling in test-time compute, 2025.
\newblock URL \url{https://arxiv.org/abs/2507.14417}.

\bibitem[ggerganov(2025)]{ggerganovllama.cpp}
ggerganov.
\newblock llama.cpp: Llm inference in c/c++.
\newblock \url{https://github.com/ggerganov/llama.cpp}, 2025.

\bibitem[Grattafiori et~al.(2024)Grattafiori, Dubey, Jauhri, Pandey, Kadian, Al-Dahle, Letman, Mathur, Schelten, Vaughan, Yang, Fan, Goyal, Hartshorn, Yang, Mitra, Sravankumar, Korenev, Hinsvark, Rao, Zhang, Rodriguez, Gregerson, Spataru, Roziere, Biron, Tang, Chern, Caucheteux, Nayak, Bi, Marra, McConnell, Keller, Touret, Wu, Wong, Ferrer, Nikolaidis, Allonsius, Song, Pintz, Livshits, Wyatt, Esiobu, Choudhary, Mahajan, Garcia-Olano, Perino, Hupkes, Lakomkin, AlBadawy, Lobanova, Dinan, Smith, Radenovic, Guzmán, Zhang, Synnaeve, Lee, Anderson, Thattai, Nail, Mialon, Pang, Cucurell, Nguyen, Korevaar, Xu, Touvron, Zarov, Ibarra, Kloumann, Misra, Evtimov, Zhang, Copet, Lee, Geffert, Vranes, Park, Mahadeokar, Shah, van~der Linde, Billock, Hong, Lee, Fu, Chi, Huang, Liu, Wang, Yu, Bitton, Spisak, Park, Rocca, Johnstun, Saxe, Jia, Alwala, Prasad, Upasani, Plawiak, Li, Heafield, Stone, El-Arini, Iyer, Malik, Chiu, Bhalla, Lakhotia, Rantala-Yeary, van~der Maaten, Chen, Tan, Jenkins, Martin, Madaan, Malo, Blecher,
  Landzaat, de~Oliveira, Muzzi, Pasupuleti, Singh, Paluri, Kardas, Tsimpoukelli, Oldham, Rita, Pavlova, Kambadur, Lewis, Si, Singh, Hassan, Goyal, Torabi, Bashlykov, Bogoychev, Chatterji, Zhang, Duchenne, Çelebi, Alrassy, Zhang, Li, Vasic, Weng, Bhargava, Dubal, Krishnan, Koura, Xu, He, Dong, Srinivasan, Ganapathy, Calderer, Cabral, Stojnic, Raileanu, Maheswari, Girdhar, Patel, Sauvestre, Polidoro, Sumbaly, Taylor, Silva, Hou, Wang, Hosseini, Chennabasappa, Singh, Bell, Kim, Edunov, Nie, Narang, Raparthy, Shen, Wan, Bhosale, Zhang, Vandenhende, Batra, Whitman, Sootla, Collot, Gururangan, Borodinsky, Herman, Fowler, Sheasha, Georgiou, Scialom, Speckbacher, Mihaylov, Xiao, Karn, Goswami, Gupta, Ramanathan, Kerkez, Gonguet, Do, Vogeti, Albiero, Petrovic, Chu, Xiong, Fu, Meers, Martinet, Wang, Wang, Tan, Xia, Xie, Jia, Wang, Goldschlag, Gaur, Babaei, Wen, Song, Zhang, Li, Mao, Coudert, Yan, Chen, Papakipos, Singh, Srivastava, Jain, Kelsey, Shajnfeld, Gangidi, Victoria, Goldstand, Menon, Sharma, Boesenberg,
  Baevski, Feinstein, Kallet, Sangani, Teo, Yunus, Lupu, Alvarado, Caples, Gu, Ho, Poulton, Ryan, Ramchandani, Dong, Franco, Goyal, Saraf, Chowdhury, Gabriel, Bharambe, Eisenman, Yazdan, James, Maurer, Leonhardi, Huang, Loyd, Paola, Paranjape, Liu, Wu, Ni, Hancock, Wasti, Spence, Stojkovic, Gamido, Montalvo, Parker, Burton, Mejia, Liu, Wang, Kim, Zhou, Hu, Chu, Cai, Tindal, Feichtenhofer, Gao, Civin, Beaty, Kreymer, Li, Adkins, Xu, Testuggine, David, Parikh, Liskovich, Foss, Wang, Le, Holland, Dowling, Jamil, Montgomery, Presani, Hahn, Wood, Le, Brinkman, Arcaute, Dunbar, Smothers, Sun, Kreuk, Tian, Kokkinos, Ozgenel, Caggioni, Kanayet, Seide, Florez, Schwarz, Badeer, Swee, Halpern, Herman, Sizov, Guangyi, Zhang, Lakshminarayanan, Inan, Shojanazeri, Zou, Wang, Zha, Habeeb, Rudolph, Suk, Aspegren, Goldman, Zhan, Damlaj, Molybog, Tufanov, Leontiadis, Veliche, Gat, Weissman, Geboski, Kohli, Lam, Asher, Gaya, Marcus, Tang, Chan, Zhen, Reizenstein, Teboul, Zhong, Jin, Yang, Cummings, Carvill, Shepard, McPhie,
  Torres, Ginsburg, Wang, Wu, U, Saxena, Khandelwal, Zand, Matosich, Veeraraghavan, Michelena, Li, Jagadeesh, Huang, Chawla, Huang, Chen, Garg, A, Silva, Bell, Zhang, Guo, Yu, Moshkovich, Wehrstedt, Khabsa, Avalani, Bhatt, Mankus, Hasson, Lennie, Reso, Groshev, Naumov, Lathi, Keneally, Liu, Seltzer, Valko, Restrepo, Patel, Vyatskov, Samvelyan, Clark, Macey, Wang, Hermoso, Metanat, Rastegari, Bansal, Santhanam, Parks, White, Bawa, Singhal, Egebo, Usunier, Mehta, Laptev, Dong, Cheng, Chernoguz, Hart, Salpekar, Kalinli, Kent, Parekh, Saab, Balaji, Rittner, Bontrager, Roux, Dollar, Zvyagina, Ratanchandani, Yuvraj, Liang, Alao, Rodriguez, Ayub, Murthy, Nayani, Mitra, Parthasarathy, Li, Hogan, Battey, Wang, Howes, Rinott, Mehta, Siby, Bondu, Datta, Chugh, Hunt, Dhillon, Sidorov, Pan, Mahajan, Verma, Yamamoto, Ramaswamy, Lindsay, Lindsay, Feng, Lin, Zha, Patil, Shankar, Zhang, Zhang, Wang, Agarwal, Sajuyigbe, Chintala, Max, Chen, Kehoe, Satterfield, Govindaprasad, Gupta, Deng, Cho, Virk, Subramanian, Choudhury,
  Goldman, Remez, Glaser, Best, Koehler, Robinson, Li, Zhang, Matthews, Chou, Shaked, Vontimitta, Ajayi, Montanez, Mohan, Kumar, Mangla, Ionescu, Poenaru, Mihailescu, Ivanov, Li, Wang, Jiang, Bouaziz, Constable, Tang, Wu, Wang, Wu, Gao, Kleinman, Chen, Hu, Jia, Qi, Li, Zhang, Zhang, Adi, Nam, Yu, Wang, Zhao, Hao, Qian, Li, He, Rait, DeVito, Rosnbrick, Wen, Yang, Zhao, and Ma]{grattafiori2024llama3herdmodels}
Grattafiori, A., Dubey, A., Jauhri, A., Pandey, A., Kadian, A., Al-Dahle, A., Letman, A., Mathur, A., Schelten, A., Vaughan, A., Yang, A., Fan, A., Goyal, A., Hartshorn, A., Yang, A., Mitra, A., Sravankumar, A., Korenev, A., Hinsvark, A., Rao, A., Zhang, A., Rodriguez, A., Gregerson, A., Spataru, A., Roziere, B., Biron, B., Tang, B., Chern, B., Caucheteux, C., Nayak, C., Bi, C., Marra, C., McConnell, C., Keller, C., Touret, C., Wu, C., Wong, C., Ferrer, C.~C., Nikolaidis, C., Allonsius, D., Song, D., Pintz, D., Livshits, D., Wyatt, D., Esiobu, D., Choudhary, D., Mahajan, D., Garcia-Olano, D., Perino, D., Hupkes, D., Lakomkin, E., AlBadawy, E., Lobanova, E., Dinan, E., Smith, E.~M., Radenovic, F., Guzmán, F., Zhang, F., Synnaeve, G., Lee, G., Anderson, G.~L., Thattai, G., Nail, G., Mialon, G., Pang, G., Cucurell, G., Nguyen, H., Korevaar, H., Xu, H., Touvron, H., Zarov, I., Ibarra, I.~A., Kloumann, I., Misra, I., Evtimov, I., Zhang, J., Copet, J., Lee, J., Geffert, J., Vranes, J., Park, J., Mahadeokar, J.,
  Shah, J., van~der Linde, J., Billock, J., Hong, J., Lee, J., Fu, J., Chi, J., Huang, J., Liu, J., Wang, J., Yu, J., Bitton, J., Spisak, J., Park, J., Rocca, J., Johnstun, J., Saxe, J., Jia, J., Alwala, K.~V., Prasad, K., Upasani, K., Plawiak, K., Li, K., Heafield, K., Stone, K., El-Arini, K., Iyer, K., Malik, K., Chiu, K., Bhalla, K., Lakhotia, K., Rantala-Yeary, L., van~der Maaten, L., Chen, L., Tan, L., Jenkins, L., Martin, L., Madaan, L., Malo, L., Blecher, L., Landzaat, L., de~Oliveira, L., Muzzi, M., Pasupuleti, M., Singh, M., Paluri, M., Kardas, M., Tsimpoukelli, M., Oldham, M., Rita, M., Pavlova, M., Kambadur, M., Lewis, M., Si, M., Singh, M.~K., Hassan, M., Goyal, N., Torabi, N., Bashlykov, N., Bogoychev, N., Chatterji, N., Zhang, N., Duchenne, O., Çelebi, O., Alrassy, P., Zhang, P., Li, P., Vasic, P., Weng, P., Bhargava, P., Dubal, P., Krishnan, P., Koura, P.~S., Xu, P., He, Q., Dong, Q., Srinivasan, R., Ganapathy, R., Calderer, R., Cabral, R.~S., Stojnic, R., Raileanu, R., Maheswari, R., Girdhar,
  R., Patel, R., Sauvestre, R., Polidoro, R., Sumbaly, R., Taylor, R., Silva, R., Hou, R., Wang, R., Hosseini, S., Chennabasappa, S., Singh, S., Bell, S., Kim, S.~S., Edunov, S., Nie, S., Narang, S., Raparthy, S., Shen, S., Wan, S., Bhosale, S., Zhang, S., Vandenhende, S., Batra, S., Whitman, S., Sootla, S., Collot, S., Gururangan, S., Borodinsky, S., Herman, T., Fowler, T., Sheasha, T., Georgiou, T., Scialom, T., Speckbacher, T., Mihaylov, T., Xiao, T., Karn, U., Goswami, V., Gupta, V., Ramanathan, V., Kerkez, V., Gonguet, V., Do, V., Vogeti, V., Albiero, V., Petrovic, V., Chu, W., Xiong, W., Fu, W., Meers, W., Martinet, X., Wang, X., Wang, X., Tan, X.~E., Xia, X., Xie, X., Jia, X., Wang, X., Goldschlag, Y., Gaur, Y., Babaei, Y., Wen, Y., Song, Y., Zhang, Y., Li, Y., Mao, Y., Coudert, Z.~D., Yan, Z., Chen, Z., Papakipos, Z., Singh, A., Srivastava, A., Jain, A., Kelsey, A., Shajnfeld, A., Gangidi, A., Victoria, A., Goldstand, A., Menon, A., Sharma, A., Boesenberg, A., Baevski, A., Feinstein, A., Kallet, A.,
  Sangani, A., Teo, A., Yunus, A., Lupu, A., Alvarado, A., Caples, A., Gu, A., Ho, A., Poulton, A., Ryan, A., Ramchandani, A., Dong, A., Franco, A., Goyal, A., Saraf, A., Chowdhury, A., Gabriel, A., Bharambe, A., Eisenman, A., Yazdan, A., James, B., Maurer, B., Leonhardi, B., Huang, B., Loyd, B., Paola, B.~D., Paranjape, B., Liu, B., Wu, B., Ni, B., Hancock, B., Wasti, B., Spence, B., Stojkovic, B., Gamido, B., Montalvo, B., Parker, C., Burton, C., Mejia, C., Liu, C., Wang, C., Kim, C., Zhou, C., Hu, C., Chu, C.-H., Cai, C., Tindal, C., Feichtenhofer, C., Gao, C., Civin, D., Beaty, D., Kreymer, D., Li, D., Adkins, D., Xu, D., Testuggine, D., David, D., Parikh, D., Liskovich, D., Foss, D., Wang, D., Le, D., Holland, D., Dowling, E., Jamil, E., Montgomery, E., Presani, E., Hahn, E., Wood, E., Le, E.-T., Brinkman, E., Arcaute, E., Dunbar, E., Smothers, E., Sun, F., Kreuk, F., Tian, F., Kokkinos, F., Ozgenel, F., Caggioni, F., Kanayet, F., Seide, F., Florez, G.~M., Schwarz, G., Badeer, G., Swee, G., Halpern, G.,
  Herman, G., Sizov, G., Guangyi, Zhang, Lakshminarayanan, G., Inan, H., Shojanazeri, H., Zou, H., Wang, H., Zha, H., Habeeb, H., Rudolph, H., Suk, H., Aspegren, H., Goldman, H., Zhan, H., Damlaj, I., Molybog, I., Tufanov, I., Leontiadis, I., Veliche, I.-E., Gat, I., Weissman, J., Geboski, J., Kohli, J., Lam, J., Asher, J., Gaya, J.-B., Marcus, J., Tang, J., Chan, J., Zhen, J., Reizenstein, J., Teboul, J., Zhong, J., Jin, J., Yang, J., Cummings, J., Carvill, J., Shepard, J., McPhie, J., Torres, J., Ginsburg, J., Wang, J., Wu, K., U, K.~H., Saxena, K., Khandelwal, K., Zand, K., Matosich, K., Veeraraghavan, K., Michelena, K., Li, K., Jagadeesh, K., Huang, K., Chawla, K., Huang, K., Chen, L., Garg, L., A, L., Silva, L., Bell, L., Zhang, L., Guo, L., Yu, L., Moshkovich, L., Wehrstedt, L., Khabsa, M., Avalani, M., Bhatt, M., Mankus, M., Hasson, M., Lennie, M., Reso, M., Groshev, M., Naumov, M., Lathi, M., Keneally, M., Liu, M., Seltzer, M.~L., Valko, M., Restrepo, M., Patel, M., Vyatskov, M., Samvelyan, M., Clark,
  M., Macey, M., Wang, M., Hermoso, M.~J., Metanat, M., Rastegari, M., Bansal, M., Santhanam, N., Parks, N., White, N., Bawa, N., Singhal, N., Egebo, N., Usunier, N., Mehta, N., Laptev, N.~P., Dong, N., Cheng, N., Chernoguz, O., Hart, O., Salpekar, O., Kalinli, O., Kent, P., Parekh, P., Saab, P., Balaji, P., Rittner, P., Bontrager, P., Roux, P., Dollar, P., Zvyagina, P., Ratanchandani, P., Yuvraj, P., Liang, Q., Alao, R., Rodriguez, R., Ayub, R., Murthy, R., Nayani, R., Mitra, R., Parthasarathy, R., Li, R., Hogan, R., Battey, R., Wang, R., Howes, R., Rinott, R., Mehta, S., Siby, S., Bondu, S.~J., Datta, S., Chugh, S., Hunt, S., Dhillon, S., Sidorov, S., Pan, S., Mahajan, S., Verma, S., Yamamoto, S., Ramaswamy, S., Lindsay, S., Lindsay, S., Feng, S., Lin, S., Zha, S.~C., Patil, S., Shankar, S., Zhang, S., Zhang, S., Wang, S., Agarwal, S., Sajuyigbe, S., Chintala, S., Max, S., Chen, S., Kehoe, S., Satterfield, S., Govindaprasad, S., Gupta, S., Deng, S., Cho, S., Virk, S., Subramanian, S., Choudhury, S.,
  Goldman, S., Remez, T., Glaser, T., Best, T., Koehler, T., Robinson, T., Li, T., Zhang, T., Matthews, T., Chou, T., Shaked, T., Vontimitta, V., Ajayi, V., Montanez, V., Mohan, V., Kumar, V.~S., Mangla, V., Ionescu, V., Poenaru, V., Mihailescu, V.~T., Ivanov, V., Li, W., Wang, W., Jiang, W., Bouaziz, W., Constable, W., Tang, X., Wu, X., Wang, X., Wu, X., Gao, X., Kleinman, Y., Chen, Y., Hu, Y., Jia, Y., Qi, Y., Li, Y., Zhang, Y., Zhang, Y., Adi, Y., Nam, Y., Yu, Wang, Zhao, Y., Hao, Y., Qian, Y., Li, Y., He, Y., Rait, Z., DeVito, Z., Rosnbrick, Z., Wen, Z., Yang, Z., Zhao, Z., and Ma, Z.
\newblock The llama 3 herd of models, 2024.
\newblock URL \url{https://arxiv.org/abs/2407.21783}.

\bibitem[Hallapy et~al.(2023)Hallapy, Hawkins, Kelley, O'Brien, and Zipkin]{milglue}
Hallapy, J., Hawkins, T., Kelley, T., O'Brien, C., and Zipkin, J.~R.
\newblock Milglue: A multitask benchmark platform for natural language understanding in the military domain.
\newblock \emph{Military Operations Research}, 28\penalty0 (1):\penalty0 pp. 97--116, 2023.
\newblock ISSN 10825983, 21632758.
\newblock URL \url{https://www.jstor.org/stable/27207617}.

\bibitem[Hassid et~al.(2025)Hassid, Synnaeve, Adi, and Schwartz]{hassid2025dontoverthinkitpreferring}
Hassid, M., Synnaeve, G., Adi, Y., and Schwartz, R.
\newblock Don't overthink it. preferring shorter thinking chains for improved llm reasoning, 2025.
\newblock URL \url{https://arxiv.org/abs/2505.17813}.

\bibitem[Hendrycks et~al.(2021)Hendrycks, Burns, Basart, Zou, Mazeika, Song, and Steinhardt]{hendryckstest2021}
Hendrycks, D., Burns, C., Basart, S., Zou, A., Mazeika, M., Song, D., and Steinhardt, J.
\newblock Measuring massive multitask language understanding.
\newblock \emph{Proceedings of the International Conference on Learning Representations (ICLR)}, 2021.

\bibitem[Jakovich(2024)]{jakovich2024quantifying}
Jakovich, R.~C.
\newblock Quantifying resilience and redundancy for air force linear infrastructure.
\newblock Master's thesis, Air Force Institute of Technology, 2024.

\bibitem[Kwon et~al.(2023)Kwon, Li, Zhuang, Sheng, Zheng, Yu, Gonzalez, Zhang, and Stoica]{kwon2023efficient}
Kwon, W., Li, Z., Zhuang, S., Sheng, Y., Zheng, L., Yu, C.~H., Gonzalez, J.~E., Zhang, H., and Stoica, I.
\newblock Efficient memory management for large language model serving with pagedattention.
\newblock In \emph{Proceedings of the ACM SIGOPS 29th Symposium on Operating Systems Principles}, 2023.

\bibitem[Lee et~al.(2023)Lee, Bubeck, and Petro]{lee2023benefits}
Lee, P., Bubeck, S., and Petro, J.
\newblock Benefits, limits, and risks of gpt-4 as an ai chatbot for medicine.
\newblock \emph{New England Journal of Medicine}, 388\penalty0 (13):\penalty0 1233--1239, 2023.

\bibitem[Lin et~al.(2022)Lin, Hilton, and Evans]{lin-etal-2022-truthfulqa}
Lin, S., Hilton, J., and Evans, O.
\newblock {T}ruthful{QA}: Measuring how models mimic human falsehoods.
\newblock In Muresan, S., Nakov, P., and Villavicencio, A. (eds.), \emph{Proceedings of the 60th Annual Meeting of the Association for Computational Linguistics (Volume 1: Long Papers)}, pp.\  3214--3252, Dublin, Ireland, May 2022. Association for Computational Linguistics.
\newblock \doi{10.18653/v1/2022.acl-long.229}.
\newblock URL \url{https://aclanthology.org/2022.acl-long.229/}.

\bibitem[Nvidia(2025)]{modelopt}
Nvidia.
\newblock Nvidia tensorrt model optimizer, 2025.
\newblock URL \url{https://github.com/NVIDIA/TensorRT-Model-Optimizer}.

\bibitem[OpenAI et~al.(2025)OpenAI, :, Agarwal, Ahmad, Ai, Altman, Applebaum, Arbus, Arora, Bai, Baker, Bao, Barak, Bennett, Bertao, Brett, Brevdo, Brockman, Bubeck, Chang, Chen, Chen, Cheung, Clark, Cook, Dukhan, Dvorak, Fives, Fomenko, Garipov, Georgiev, Glaese, Gogineni, Goucher, Gross, Guzman, Hallman, Hehir, Heidecke, Helyar, Hu, Huet, Huh, Jain, Johnson, Koch, Kofman, Kundel, Kwon, Kyrylov, Le, Leclerc, Lennon, Lessans, Lezcano-Casado, Li, Li, Lin, Liss, Lily, Liu, Liu, Lu, Lu, Martinovic, McCallum, McGrath, McKinney, McLaughlin, Mei, Mostovoy, Mu, Myles, Neitz, Nichol, Pachocki, Paino, Palmie, Pantuliano, Parascandolo, Park, Pathak, Paz, Peran, Pimenov, Pokrass, Proehl, Qiu, Raila, Raso, Ren, Richardson, Robinson, Rotsted, Salman, Sanjeev, Schwarzer, Sculley, Sikchi, Simon, Singhal, Song, Stuckey, Sun, Tillet, Toizer, Tsimpourlas, Vyas, Wallace, Wang, Wang, Watkins, Weil, Wendling, Whinnery, Whitney, Wong, Yang, Yang, Yasunaga, Ying, Zaremba, Zhan, Zhang, Zhang, Zhang, and
  Zhao]{openai2025gptoss120bgptoss20bmodel}
OpenAI, :, Agarwal, S., Ahmad, L., Ai, J., Altman, S., Applebaum, A., Arbus, E., Arora, R.~K., Bai, Y., Baker, B., Bao, H., Barak, B., Bennett, A., Bertao, T., Brett, N., Brevdo, E., Brockman, G., Bubeck, S., Chang, C., Chen, K., Chen, M., Cheung, E., Clark, A., Cook, D., Dukhan, M., Dvorak, C., Fives, K., Fomenko, V., Garipov, T., Georgiev, K., Glaese, M., Gogineni, T., Goucher, A., Gross, L., Guzman, K.~G., Hallman, J., Hehir, J., Heidecke, J., Helyar, A., Hu, H., Huet, R., Huh, J., Jain, S., Johnson, Z., Koch, C., Kofman, I., Kundel, D., Kwon, J., Kyrylov, V., Le, E.~Y., Leclerc, G., Lennon, J.~P., Lessans, S., Lezcano-Casado, M., Li, Y., Li, Z., Lin, J., Liss, J., Lily, Liu, Liu, J., Lu, K., Lu, C., Martinovic, Z., McCallum, L., McGrath, J., McKinney, S., McLaughlin, A., Mei, S., Mostovoy, S., Mu, T., Myles, G., Neitz, A., Nichol, A., Pachocki, J., Paino, A., Palmie, D., Pantuliano, A., Parascandolo, G., Park, J., Pathak, L., Paz, C., Peran, L., Pimenov, D., Pokrass, M., Proehl, E., Qiu, H., Raila, G.,
  Raso, F., Ren, H., Richardson, K., Robinson, D., Rotsted, B., Salman, H., Sanjeev, S., Schwarzer, M., Sculley, D., Sikchi, H., Simon, K., Singhal, K., Song, Y., Stuckey, D., Sun, Z., Tillet, P., Toizer, S., Tsimpourlas, F., Vyas, N., Wallace, E., Wang, X., Wang, M., Watkins, O., Weil, K., Wendling, A., Whinnery, K., Whitney, C., Wong, H., Yang, L., Yang, Y., Yasunaga, M., Ying, K., Zaremba, W., Zhan, W., Zhang, C., Zhang, B., Zhang, E., and Zhao, S.
\newblock gpt-oss-120b \& gpt-oss-20b model card, 2025.
\newblock URL \url{https://arxiv.org/abs/2508.10925}.

\bibitem[Pal et~al.(2022)Pal, Umapathi, and Sankarasubbu]{pmlr-v174-pal22a}
Pal, A., Umapathi, L.~K., and Sankarasubbu, M.
\newblock Medmcqa: A large-scale multi-subject multi-choice dataset for medical domain question answering.
\newblock In \emph{Proceedings of the Conference on Health, Inference, and Learning}, volume 174 of \emph{Proceedings of Machine Learning Research}, pp.\  248--260. PMLR, 07--08 Apr 2022.
\newblock URL \url{https://proceedings.mlr.press/v174/pal22a.html}.

\bibitem[Parham \& Lynn(2025)Parham and Lynn]{GovBench:JointStaffBench}
Parham, G. and Lynn, J.
\newblock Jointstaffbench, 2025.
\newblock URL \url{https://www.govbench.ai/jointstaffbench}.
\newblock Accessed: October 2025.

\bibitem[Rein et~al.(2023)Rein, Hou, Stickland, Petty, Pang, Dirani, Michael, and Bowman]{rein2023gpqagraduatelevelgoogleproofqa}
Rein, D., Hou, B.~L., Stickland, A.~C., Petty, J., Pang, R.~Y., Dirani, J., Michael, J., and Bowman, S.~R.
\newblock Gpqa: A graduate-level google-proof q\&a benchmark, 2023.
\newblock URL \url{https://arxiv.org/abs/2311.12022}.

\bibitem[{Reuters}(2025)]{reuters_openai_2025}
{Reuters}.
\newblock {OpenAI} hits \$12 billion in annualized revenue, information reports, jul 2025.

\bibitem[Ruiz \& Sell(2024)Ruiz and Sell]{ruiz2024fine}
Ruiz, D.~C. and Sell, J.
\newblock Fine-tuning and evaluating open-source large language models for the army domain.
\newblock \emph{arXiv preprint arXiv:2410.20297}, 2024.

\bibitem[{UK Government BEIS}(2025)]{inspect_evals}
{UK Government BEIS}.
\newblock {inspect\_evals}: Collection of evals for inspect ai, 2025.
\newblock URL \url{https://github.com/UKGovernmentBEIS/inspect_evals}.
\newblock Accessed: 6 Sep 2025.

\bibitem[von Werra et~al.(2020)von Werra, Belkada, Tunstall, Beeching, Thrush, Lambert, Huang, Rasul, and Gallouédec]{vonwerra2022trl}
von Werra, L., Belkada, Y., Tunstall, L., Beeching, E., Thrush, T., Lambert, N., Huang, S., Rasul, K., and Gallouédec, Q.
\newblock Trl: Transformer reinforcement learning.
\newblock \url{https://github.com/huggingface/trl}, 2020.

\bibitem[Wang et~al.(2025{\natexlab{a}})Wang, Gong, Zhang, Xu, and Wang]{wang2025aiagenticprogrammingsurvey}
Wang, H., Gong, J., Zhang, H., Xu, J., and Wang, Z.
\newblock Ai agentic programming: A survey of techniques, challenges, and opportunities, 2025{\natexlab{a}}.
\newblock URL \url{https://arxiv.org/abs/2508.11126}.

\bibitem[Wang et~al.(2025{\natexlab{b}})Wang, Ma, Wang, Wu, Ji, Chen, Li, and Yuan]{wang2025survey}
Wang, W., Ma, Z., Wang, Z., Wu, C., Ji, J., Chen, W., Li, X., and Yuan, Y.
\newblock A survey of llm-based agents in medicine: How far are we from baymax?
\newblock \emph{arXiv preprint arXiv:2502.11211}, 2025{\natexlab{b}}.

\bibitem[Wang et~al.(2024)Wang, Ma, Zhang, Ni, Chandra, Guo, Ren, Arulraj, He, Jiang, Li, Ku, Wang, Zhuang, Fan, Yue, and Chen]{wang2024mmluprorobustchallengingmultitask}
Wang, Y., Ma, X., Zhang, G., Ni, Y., Chandra, A., Guo, S., Ren, W., Arulraj, A., He, X., Jiang, Z., Li, T., Ku, M., Wang, K., Zhuang, A., Fan, R., Yue, X., and Chen, W.
\newblock Mmlu-pro: A more robust and challenging multi-task language understanding benchmark, 2024.
\newblock URL \url{https://arxiv.org/abs/2406.01574}.

\bibitem[Wu et~al.(2025)Wu, Liu, Zhang, Chen, Meng, Du, Zhao, Zhang, Ye, Wang, Wang, Ni, Yang, Xu, and Shieh]{wu2025mcpmarkbenchmarkstresstestingrealistic}
Wu, Z., Liu, X., Zhang, X., Chen, L., Meng, F., Du, L., Zhao, Y., Zhang, F., Ye, Y., Wang, J., Wang, Z., Ni, J., Yang, Y., Xu, A., and Shieh, M.~Q.
\newblock Mcpmark: A benchmark for stress-testing realistic and comprehensive mcp use, 2025.
\newblock URL \url{https://arxiv.org/abs/2509.24002}.

\bibitem[Yoo et~al.(2003)Yoo, Jette, and Grondona]{slurm}
Yoo, A.~B., Jette, M.~A., and Grondona, M.
\newblock Slurm: Simple linux utility for resource management.
\newblock In Feitelson, D., Rudolph, L., and Schwiegelshohn, U. (eds.), \emph{Job Scheduling Strategies for Parallel Processing}, pp.\  44--60, Berlin, Heidelberg, 2003. Springer Berlin Heidelberg.
\newblock ISBN 978-3-540-39727-4.

\bibitem[Zhang et~al.(2024)Zhang, Chen, Jin, Wang, Ji, Wang, and Han]{zhang2024comprehensive}
Zhang, Y., Chen, X., Jin, B., Wang, S., Ji, S., Wang, W., and Han, J.
\newblock A comprehensive survey of scientific large language models and their applications in scientific discovery.
\newblock In \emph{Proceedings of the 2024 Conference on Empirical Methods in Natural Language Processing}, pp.\  8783--8817, 2024.

\bibitem[Zhou et~al.(2023)Zhou, Lu, Mishra, Brahma, Basu, Luan, Zhou, and Hou]{zhou2023instructionfollowingevaluationlargelanguage}
Zhou, J., Lu, T., Mishra, S., Brahma, S., Basu, S., Luan, Y., Zhou, D., and Hou, L.
\newblock Instruction-following evaluation for large language models, 2023.
\newblock URL \url{https://arxiv.org/abs/2311.07911}.

\end{thebibliography}
\bibliographystyle{mlsys2025}

\clearpage
\appendix
\section{Test Set Examples}

Examples from our four test sets are provided in Table \ref{tab:test-set-examples}.

\begin{table*}[htbp]
\caption[]{Samples from each of our four tests.}
\label{tab:test-set-examples}
\vspace{3pt}
\centering
\small
\begin{tblr}{
    colspec={X[1,l]X[2,j]X[4,j]},
    rows={t},
    row{1}={font=\bfseries}
}
\hline[1pt]
Test Set & Question & Answer \\
\hline[0.5pt]
\textsc{combat-arms} & How do artillery units provide counterfire support in a contested EW environment? & By using preplanned fire missions, manual gunnery solutions, redundant comms, and passive sensors to detect enemy artillery. \\
\textsc{combat-medic} & According to ATP 4-25.12 (Unit Field Sanitation Team), what are the steps a 68W Combat Medic should take to ensure proper disease prevention measures in a deployed environment? & Under ATP 4-25.12, a 68W Combat Medic should assess the operational environment, implement unit-level field sanitation measures for disease prevention measures, coordinate with the Unit Field Sanitation Team, and provide preventive medicine guidance to reduce disease and non-battle injuries, ensuring soldier readiness. \\
\textsc{cyber} & Explain how ethical hacking engagements strengthen organizational resilience. & Ethical hacking involves authorized testing to identify weaknesses before adversaries can exploit them. Results inform patch management, training, and policy revisions. By exploiting gaps safely, ethical hackers validate incident response capabilities and measure defense depth. Regular testing fosters a security culture and continuous improvement loop. \\
\textsc{mil-bench-5k} & How does the concept of ``systemic disruption'' contribute to defeating adversaries in urban operations, and what are the key mechanisms involved? & {Systemic disruption serves as a decisive strategy in urban operations by degrading an adversary's capacity to function as an integrated force within complex urban environments. This approach prioritizes:\\1. **Disruption of Support Systems**\\- Isolating adversaries from critical resources (supplies, communication networks, financial infrastructure)\\- Severing ties to local support networks through population engagement and counterintelligence\\2. **Precision Force Application**\\- Targeted use of kinetic and non-kinetic effects to degrade adversary capabilities while minimizing infrastructure damage\\- Leveraging advanced ISR (Intelligence, Surveillance, Reconnaissance) for high-value target identification\\3. **Population-Centric Operations**\\- Building trust with local populations to deny adversary recruitment and logistical access\\- Delivering essential services to establish legitimacy and counter adversary narratives\\By combining these mechanisms, joint forces can collapse adversarial influence while preserving urban systems' functionality, aligning with the principles of proportionality and military necessity in densely populated areas.} \\
\hline[1pt]
\end{tblr}
\end{table*}

\section{Training Setup}
\label{sect:app-training-setup}

The alpaca chat template is provided in Figure \ref{fig:alpaca}.

\begin{figure*}[htb]
\caption{The alpaca chat template.}
\label{fig:alpaca}
\begin{lstlisting}[breaklines=true, breakautoindent=false]
SYSTEM: Below is an instruction from a USER that describes a task. The ASSISTANT writes a response that appropriately and concisely completes the request.
USER: {question}
ASSISTANT: {answer}
\end{lstlisting}
\end{figure*}

\section{Evaluation Setup}
\label{sect:app-eval-setup}

\begin{figure*}[htb]
\caption{An approximated version of the task definition used for all four military test sets.}
\label{fig:task-def}
\footnotesize
\begin{lstlisting}[language=Python]
from inspect_ai import Task, task
from inspect_ai.dataset import FieldSpec, hf_dataset
from inspect_ai.model import get_model
from inspect_ai.scorer import model_graded_qa
from inspect_ai.solver import generate

@task
def some_military_task(
    scorer_model: str = None,
    scorer_base_url: str = None,
    scorer_model_path: str = None,
    dataset_name: str = None,
    dataset_revision: str = None,
    split: str = None,
    epochs: int = 3,
    **kwargs,
) -> Task:
        
    dataset = hf_dataset(
        dataset_name,
        revision=dataset_revision,
        split=split,
        sample_fields=FieldSpec(
            input="question",
            target="answer",
            metadata=[
                "category", "difficulty", "reference"
            ]
        ),
    )
    
    # if there is a running openai server (EX: llama.cpp)
    if scorer_base_url:
        model = get_model(model=scorer_model, base_url=scorer_base_url)
    # for vLLM
    elif scorer_model == "vllm/local" and scorer_model_path:
        model = get_model(
            # note: You need to ultimately have two slashes after vllm
            model="vllm/"+scorer_model_path,
            **kwargs,
        )
    else:
        model = scorer_model

    scorer = model_graded_qa(model=model, partial_credit=True)

    solver = [generate()]
    
    return Task(
        dataset=dataset,
        solver=solver,
        scorer=scorer,
        epochs=epochs,
    )
\end{lstlisting}
\end{figure*}

All four military test sets use the Inspect task definition code provided in Figure \ref{fig:task-def}, with some minor subtleties from task to task. The LLM Judge prompt is provided in Figure \ref{fig:judge-prompt}.

\begin{figure*}[htbp]
\caption{The default LLM Judge prompt used in Inspect, which was used for all four military tests.}
\label{fig:judge-prompt}
\begin{lstlisting}[breaklines=true]
You are assessing a submitted answer on a given task based on a criterion. Here is the data:

[BEGIN DATA]
***
[Task]: {question}
***
[Submission]: {answer}
***
[Criterion]: {criterion}
***
[END DATA]

Does the submission meet the criterion?

After assessing the submitted answer, reply with 'GRADE: $LETTER' (without quotes) where LETTER is one of CPI. Please choose ONE option for the grade: either "C" for correct answers, "P" for partially correct answers, or "I" for incorrect answers.

For example, after reviewing a correct answer you might write 'GRADE: C' or after reviewing an incorrect answer you might write 'GRADE: I'.

First, write out in a step by step manner your reasoning about the criterion to be sure that your conclusion is correct. Avoid simply stating the correct answers at the outset. Then, end with your answer formatted as 'GRADE: $LETTER' (without quotes) where LETTER is one of CPI.
\end{lstlisting}
\end{figure*}

\end{document}